%% file: main.tex
\documentclass[acmsmall,screen]{acmart}
\AtBeginDocument{%
  \providecommand\BibTeX{{%
    \normalfont B\kern-0.5em{\scshape i\kern-0.25em b}\kern-0.8em\TeX}}}

\setcopyright{acmcopyright}
\copyrightyear{2024}
\acmYear{2024}
\acmDOI{XXXXXXX.XXXXXXX}

\acmConference[CSCW '24]{ACM Conference On Computer-Supported Cooperative Work And Social Computing}{November 9--13, 2024}{San José, Costa Rica}
\acmPrice{15.00}
\acmISBN{978-1-4503-XXXX-X/18/06}

\usepackage{listings}
\usepackage{subcaption} 
\usepackage{longtable} %
\usepackage{enumitem}
\usepackage{multicol}

\setcopyright{none}
\renewcommand\footnotetextcopyrightpermission[1]{} % removes footnote with conference information in first column
\begin{document}

\title[Mindful Explanations]{Mindful Explanations: Prevalence and Impact of Mind Attribution in XAI Research}

\author{Susanne Hindennach}
\affiliation{\institution{University of Stuttgart}\country{Germany}}\orcid{https://orcid.org/0000-0001-5072-0514}
\email{susanne.hindennach@vis.uni-stuttgart.de}

\author{Lei Shi}
\affiliation{\institution{University of Stuttgart}\country{Germany}}
\email{lei.shi@vis.uni-stuttgart.de}
\orcid{https://orcid.org/0000-0003-1628-1559}

\author{Filip Miletić}
\affiliation{\institution{University of Stuttgart}\country{Germany}}
\email{filip.miletic@ims.uni-stuttgart.de}
\orcid{https://orcid.org/0000-0003-1147-196X}

\author{Andreas Bulling}
\affiliation{\institution{University of Stuttgart}\country{Germany}}
\email{andreas.bulling@vis.uni-stuttgart.de}
\orcid{https://orcid.org/0000-0001-6317-7303}

\renewcommand{\shortauthors}{Hindennach, Shi, Miletić, Bulling}

\begin{abstract}
\input{abstract_new.tex}
\end{abstract}

\begin{CCSXML}
<ccs2012>
   <concept>
       <concept_id>10003120.10003121.10003126</concept_id>
       <concept_desc>Human-centered computing~HCI theory, concepts and models</concept_desc>
       <concept_significance>500</concept_significance>
       </concept>
   <concept>
       <concept_id>10003120.10003130.10003131</concept_id>
       <concept_desc>Human-centered computing~Collaborative and social computing theory, concepts and paradigms</concept_desc>
       <concept_significance>500</concept_significance>
       </concept>
   <concept>
       <concept_id>10010147.10010178.10010216.10010218</concept_id>
       <concept_desc>Computing methodologies~Theory of mind</concept_desc>
       <concept_significance>500</concept_significance>
       </concept>
 </ccs2012>
\end{CCSXML}

\ccsdesc[500]{Human-centered computing~HCI theory, concepts and models}
\ccsdesc[500]{Human-centered computing~Collaborative and social computing theory, concepts and paradigms}
\ccsdesc[500]{Computing methodologies~Theory of mind}

\keywords{}

\begin{teaserfigure}
     \centering
     \includegraphics[width=0.9\textwidth,trim={1cm 6cm 1.4cm 7cm},clip]{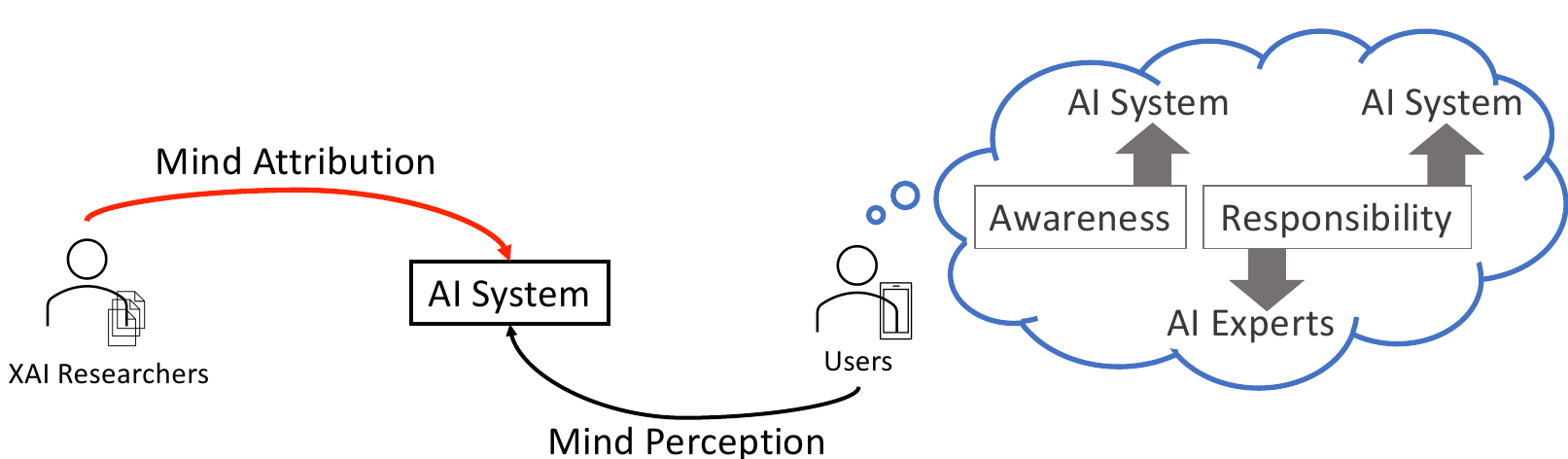}
    \caption{We study whether XAI researchers attribute a mind to AI systems by using mental verbs like ``to think'' in their explanations, which might result in users perceiving AI systems as mindful, independent agents. We investigate this impact in a vignette-based experiment and find that users rate the system's awareness higher, and consequently hold the AI system responsible even after considering the responsibility of AI experts.}
    \label{fig:concept}
\end{teaserfigure}

\maketitle

\input{introdution.tex}

\input{related_work.tex}

\input{analysis.tex}

\input{effect.tex}

\input{discussion.tex}

\begin{acks}
Susanne Hindennach and Andreas Bulling were funded by the European Research Council (ERC; grant agreement 801708).
Filip Miletić was funded by the Deutsche Forschungsgemeinschaft (DFG, German Research Foundation; research grant SCHU 2580/5-1).
Lei Shi was funded by Deutsche Forschungsgemeinschaft (DFG, German Research Foundation) under Germany's Excellence Strategy - EXC 2075 – 390740016. We acknowledge the support by the Stuttgart Center for Simulation Science (SimTech). We thank Judith Schepers for helping with the ordinal regression models. 
\end{acks}

\bibliographystyle{ACM-Reference-Format}
\bibliography{references}

\pagebreak
\appendix

\input{appendix.tex}

\end{document}

%% file: abstract_new.tex
When users perceive AI systems as mindful, independent agents, they hold them responsible instead of the AI experts who created and designed these systems. 
So far, it has not been studied whether explanations support this shift in responsibility through the use of mind-attributing verbs like ``to think''. 
To better understand the prevalence of mind-attributing explanations we analyse AI explanations in 3,533 explainable AI (XAI) research articles from the Semantic Scholar Open Research Corpus (S2ORC).
Using methods from semantic shift detection, we identify three dominant types of mind attribution: (1) metaphorical (e.g. ``to learn'' or ``to predict''), (2) awareness (e.g. ``to consider''), and (3) agency (e.g. ``to make decisions''). 
We then analyse the impact of mind-attributing explanations 
on awareness and responsibility in a vignette-based experiment with 199 participants.
We find that participants who were given a mind-attributing explanation were more likely to rate the AI system as aware of the harm it caused. 
Moreover, the mind-attributing explanation had a responsibility-concealing effect: 
Considering the AI experts' involvement lead to reduced ratings of AI responsibility for participants who were given a non-mind-attributing or no explanation. In contrast, participants who read the mind-attributing explanation still held the AI system responsible despite considering the AI experts' involvement. 
Taken together, our work underlines the need to carefully phrase explanations about AI systems in scientific writing to reduce mind attribution and clearly communicate human responsibility.  

%% file: introdution.tex
\section{Introduction}
Explanations are important to ensure accountability and to help users better understand and critically assess %
artificial intelligence (AI) systems \cite{Adadi2018,meyerResponsibilityHybridSocieties2023}.
However, users' understanding not only depends on the quality of the explanations generated by explainable AI (XAI) methods but also on the language used in these explanations.
The choice of terms, such as ``computer program'' vs. ``artificial intelligence'', can affect the perception and evaluation of the systems' abilities 
\cite{langerLookItComputer2022}.
Similarly, the usage of mental verbs like ``to think'', ``to know'', or ``to understand'' for AI systems suggests that the system is mindful and intelligent. 
We refer to the latter as \textit{mind-attributing language} (or \textit{mind attribution}
\footnote{There are other non-linguistic forms of mind attribution, such as the design choice of showing moving dots while a chatbot generates output which make it seem like the system is thinking and typing its answer like a human.} in short), i.e. the usage of terms that refer to mental states of AI systems that these systems do not necessarily have (red arrow in \autoref{fig:concept}). 

In the first part of this paper we analyse the prevalence of mind-attributing language in XAI research.
Authors and designers of XAI methods verbally attribute a mind to an AI system when they use a phrase like, for example,  ``the algorithm considers these features important'' to describe feature importance (adapted from \cite{ribeiroWhyShouldTrust2016}).
Importantly, mind attribution is different from mind perception (black arrow in \autoref{fig:concept}) that has been investigated in prior work \cite{grayDimensionsMindPerception2007, schweitzerLanguageWindowMind2021, scottYouMindUser2023}: 
Mind-attributing language can be used metaphorically without actually perceiving that the system has a mind. 
Moreover, whether or not a user perceives an AI system as mindful might depend on how the AI system is depicted (the designers' mind attribution) and the individual tendency to anthropomorphise \cite{waytzWhoSeesHuman2010}, as well as on the abilities of the AI system.

In the second part of the paper we draw on our separation and explore the influence of mind attribution on mind perception in a vignette-based experiment. Hereby, we also investigate whether mind-attributing language in explanations counteracts the goal of XAI
to maintain accountability \cite{Adadi2018} and to help users understand the constrained dependent agency of the AI system. 
Explanations might instead conceal the responsibility of the involved humans: AI systems that provide explanations of their actions might be perceived, sometimes even solely, responsible and blameworthy for their actions \cite{Lima2022}.
This is problematic because only the involved humans (AI experts, owners and users) can be held fully responsible as they define the goals of the AI system, and the amount of unsupervised actions the system can take \cite{nyholmAttributingAgencyAutomated2018}. 
We investigate whether mind attribution in explanations further amplifies this effect  by influencing mind perception which is one of the crucial reasons why AI systems are held responsible \cite{grayDimensionsMindPerception2007,shankAttributionsMoralityMind2018, voiklisMoralJudgmentsHuman2016, stuartGuiltyArtificialMinds2021}. 
More specifically, we assess the influence of mind-attributing explanations on the perception of the system's awareness of harm and whether they result in the AI system being held more responsible, and AI experts being held less responsible (blue thought bubble in \autoref{fig:concept}).

As such, the paper makes two important contributions. First, 
we use state-of-the-art methods of semantic shift detection to analyse the prevalence of mind-attributing language in XAI research. Second, we find that especially mind-attributing explanations increase the amount of awareness attributed to AI systems, 
and impede participants from reducing their rating of the AI system's responsibility after considering the involvement of AI experts.
This analysis of the prevalence and impact of mind attribution is an important step towards understanding the socio-technical implications of XAI: 
The language used by XAI researchers may suggest that AI systems are human-like agents and thereby counteracts the goal of XAI to maintain accountability.

%% file: related_work.tex
\section{Related Work}
Our work is related to previous work on shift of moral responsibility, works that have studied the link between mind attribution, mind perception and responsibility, and works on mind attribution used in (X)AI research, as well as methods for detecting mind-attributing language.

\subsection{Shift of Moral Responsibility}

Previous work %
has shown that AI systems are held responsible in the same way as a human agent would be~\cite{malleAISkyHow2019,malleWhichRobotAm2016,hongRacismResponsibilityAutonomy2019, stuartGuiltyArtificialMinds2021}. The trolley scenario, in which an agent has to decide between no action which risks the live of multiple persons or taking action and knowingly harming an individual, has been used widely as an exemplary moral decision \cite{awadMoralMachineExperiment2018, malleAISkyHow2019, malleWhichRobotAm2016, leePeopleMayPunish2021}. 
Two studies found that 60-70\% of participants blamed artificial agents in the same way they blamed human agents performing the same action in the scenario \cite{malleAISkyHow2019, malleWhichRobotAm2016}.
Another study investigated whether an AI system or a human expert was blamed differently when making a racist decision \cite{hongRacismResponsibilityAutonomy2019}. The authors found that participants held the expert and AI system equally responsible.
In addition, participants deemed the same potentially harmful action of a robot, a human, and a company equally blameworthy \cite{stuartGuiltyArtificialMinds2021}. 
Participants also rated AI systems advising or making the decision in a bail scenario similar to humans when considering causal responsibility (one of multiple investigated notions of responsibility) \cite{limaHumanPerceptionsMoral2021}.
These studies indicate that AI systems are ascribed independent human-like responsibility, however, it has not been studied yet whether mind-attributing language has an effect on this ascription.

The picture is clear when AI systems are presented in a human-like way but becomes less clear when real-world scenarios and less human-like descriptions are used. 
In the study by \citet{shankAttributionsMoralityMind2018}, participants read descriptions of seven real-world scenarios in which a moral violation occurred. 
They found that only 43.5\% of participants were sure that a moral violation occurred. %
In addition, the fact that the AI system is ``doing most of the work'', as  \citet{nyholmAttributingAgencyAutomated2018} puts it, seems to protect the more distal entities (AI experts and company), as the judgement of their action was not changed by the occurrence of a moral violation~\citep{shankAttributionsMoralityMind2018}.
The inverse outcome effect described by \citet{stuartGuiltyArtificialMinds2021} could be an indication that AI systems are implicitly held less responsible. In their study, a robot was deemed less blameworthy in the event of a bad outcome than in a neutral outcome. %
The authors hypothesised that participants were implicitly aware of the danger of shifting too much responsibility on the AI system. 
Furthermore, AI agents were assigned a lower degree of present-looking and forward-looking responsibility (the notion of responsibility as obligation, task, power and authority) than human agents \cite{limaHumanPerceptionsMoral2021}. 
So far, no study has investigated whether mentioning that the involved humans know about a potential negative outcome changes the distribution of responsibility between the human and the AI system.

\subsection{Link between Mind Attribution, Mind Perception and Responsibility}
\label{sec:MindResp}
Previous research has shown that the use of mind-attributing language is a valid measure of mind perception in realistic contexts \cite{schweitzerLanguageWindowMind2021, orrDevelopmentValidationMentalPhysical2022}. In multiple studies, \citet{schweitzerLanguageWindowMind2021} have shown that differences in the number of terms from their Mind Perception Dictionary (MPD) are representative of mind perception and downstream consequences. 
In a similar vein, \citet{orrDevelopmentValidationMentalPhysical2022} proposed Mental-Physical Verb Norms (MPVN) that measure whether a verb refers to a mental vs. physical activity. 

As a consequence, mind-attributing language might amplify the perception of a mind in AI systems, which plays a crucial role in holding an AI system responsible
\cite{grayDimensionsMindPerception2007, voiklisMoralJudgmentsHuman2016, leePeopleMayPunish2021, stuartGuiltyArtificialMinds2021, shankAttributionsMoralityMind2018}. 
Research in mind perception differentiates between two dimensions: experience and agency \cite{grayDimensionsMindPerception2007}. Experience includes capacities like hunger, fear, pain, pleasure and other emotions. Agency covers capacities like self-control, morality, memory, emotion recognition and other capabilities necessary to plan and think.  
According to \citet{grayDimensionsMindPerception2007}, responsibility is associated with high levels of agency, and experience is matched with moral patiency, i.e. being the recipient of moral wrong-doing. %
In fact, \citet{voiklisMoralJudgmentsHuman2016} showed that participants blamed a robot depending on how much they perceived it to have mental agency. 
In their study, \citet{shankAttributionsMoralityMind2018} found that the more the AI system is perceived as mindful, the more its actions are rated as wrong, and that %
including information on the AI system increased mind perception. 
Another series of three studies (two online and one lab study) found inconsistent results regarding the relationship between participants' mind perception and ratings of blame and punishment \cite{leePeopleMayPunish2021}. In all three studies, participants were presented a robot that told the trolley scenario from its first-person perspective either with or without an emotional response. 
They found that perceived patiency predicted punishment ratings (a proxy for responsibility) in the two online studies, but not in the lab study. 
\citet{stuartGuiltyArtificialMinds2021} showed that 
ratings of blame depend on whether or not an agent is attributed knowledge about a potential harm, independent of whether the agent is an AI system, a human or a company. Participants rated the knowledge of all agents similarly, and did not retrospectively prefer a formulation downgraded using quotation marks (``know'') for the AI system. 
While these works show a link between mind perception and responsibility, it has not yet been studied whether using mind-attributing formulations to first describe an AI system has an effect on the perception of mental states and consequently responsibility.

\subsection{Mind Attribution in (X)AI Research}

Mind attribution is used in (X)AI research to fulfil two communicative intents: to unify the community and to explain complex processes in simple, more tangible terms. 
The very term \textit{artificial intelligence} draws on the reference of mental states and as such, mind attribution 
is used to define and explain what AI is and should be \cite{nataleImaginingThinkingMachine2020}.
This serves to unify the research community and create momentum, but the goal to replicate mental functions has also led to controversy and heightened expectations \cite{nataleImaginingThinkingMachine2020}. 
Apart from being used to define the goal of AI research, mental state references are also used to explain AI systems without giving a detailed account of the underlying computational processes. 
In terms of the way humans make sense of behaviour, this is in line with the intentional stance defined by \citet{Dennett1971}.
He distinguished between different explanatory stances that humans can take to make sense of behaviour: 
while the physical stance explains the behaviour of a system by physical laws, the design stance leverages its purpose and known function to describe its behaviour.
Finally, the intentional stance assumes that the system will act rationally according to its mental states like desires and beliefs.
Consequently, the attribution of a mind can help to understand and predict the behaviour of an AI system, without it being the most precise representation of its underlying processes.

\subsection{Methods for Detecting Mind-Attributing Language}
The two existing measures (MPD \cite{schweitzerLanguageWindowMind2021} and MPVN \cite{orrDevelopmentValidationMentalPhysical2022}) to identify mind-attributing language 
rely on a fixed set of words whose frequency is measured in 
a given piece of text. If this selection is incomplete and other mind-attributing terms occur in the text, they would not be picked up. 
Moreover, the measures are not able to discriminate between different meanings of a word, where one meaning might attribute a mind, and the other meaning might not, as in the case of ``make'' which could refer to a physical action (``make a salad'') or a mental process (``make a guess''). %

These issues can be addressed by using computational methods to analyse word meaning, as attested in large amounts of textual data, in a bottom-up manner.
The methods draw on the principles of distributional semantics, which assumes that a word's meaning is reflected by its patterns of co-occurrence with other words in a corpus \cite{harrisDistributionalStructure1954}.
The resulting computational semantic representations range from approaches directly incorporating co-occurrence frequencies to state-of-the-art methods based on deep neural network architectures \cite{boledaDistributionalSemanticsLinguistic2020}.
Importantly, the more recent methods readily produce contextualised word embeddings, i.e. semantic representations of a word's individual occurrence, and can therefore distinguish between a word's different meanings.
This ability has been leveraged in applications such as semantic shift detection, which investigates changes in word meaning.
It often relies on clustering contextualised embeddings to automatically identify similar uses of a target word in a corpus
\cite{giulianelliAnalysingLexicalSemantic2020, montariolScalableInterpretableSemantic2021, mileticDetectingContactInducedSemantic2021}.
We thus address mind attribution in terms of detecting semantic shifts in the use of words denoting AI systems. We hypothesise that the clustering of contextualised embeddings will separate between contexts that are mind-attributing (``the model thinks'') and those that are non-mind-attributing (``the model computes'').

\subsection{Hypotheses}
In this work, we explore the following three hypotheses. 
Based on previous work on mind attribution in XAI research and the recent methods used in semantic shift detection, we propose:
\textit{There is mind-attributing language in XAI research and it can be revealed using clustering of contextualised language embeddings (Prevalence Hypothesis [$H_P$]).} 

Second, the summarised research on the link between mind attribution, mind perception and responsibility leads us to hypothesise:
\textit{ 
Mind-attributing explanations in descriptions of an AI system influence the rating of (1) awareness of the AI system, (2) responsibility of the AI system, (3) responsibility of the AI experts (Impact Hypothesis  [$H_I$]).}

Lastly, we investigate whether the shift of responsibility from humans to AI systems can be counteracted by making the human contribution explicit:
\textit{Reading that the involved humans know about a potential negative outcome and rating their responsibility decreases the rating of responsibility of the AI system (Shift Hypothesis [$H_S$]).}

%% file: analysis.tex
\section{Explanations of AI Systems in XAI Research}
To study the prevalence of mind attribution in explanations of AI systems [$H_P$], we first analysed the language in XAI research papers.
To this end we developed an analysis methodology that draws from research on semantic shift detection.
We first selected and annotated target words used to refer to AI systems in the top cited AI research articles. %
This list served as the basis for filtering the XAI research papers in the Semantic Scholar Open Research Corpus (S2ORC) for sentences containing the target words.  
We then applied state-of-the-art methods of semantic shift detection to find the common usages of the target words in the corpus.
For a selected subset of the resulting clusters, we manually classified the types of mind attribution.

\subsection{Target Word Selection}
\label{sec:target_word_selection}

The goal of this first step is to identify a set of \textit{target words} commonly used to refer to AI systems, such as ``model'' or ``neural network''.
To this end we created a list of target words used in the 12 most highly cited articles published at the top AI conferences ICLR, NeurIPS, ICML, and AAAI.
We iteratively added words until 
all potential target words in the articles were already in the target word list. 
This initial selection was annotated by three of the authors (SH, FM, LS) independently. 
Each of the annotators checked whether the candidate term was a synonym for AI system, excluding too general/specific terms, tasks, and parts of AI systems. %
The annotators agreed on 23 target words unanimously, and two out of three annotators agreed on seven target words. We kept the latter as the disagreement was caused by varying levels of expertise in subfields of AI. For example Restricted Boltzman Machines and WGAN (Wasserstein Generative Adversarial Network) were unknown to one of the annotators, respectively.
In total, this resulted in 30 target words with the most frequent ones being ``model'', ``algorithm'', and ``network''. 
See \autoref{tab:target_words} in appendix for the full list of target words. 

\subsection{Sentence Selection}
\label{sec:corpus}
We used the Semantic Scholar Open Research Corpus (S2ORC) \cite{lo-wang-2020-s2orc}. It contains parsed full-texts and abstracts from PDF and Latex sources of 12M academic papers up until April 2020. 
We filtered the corpus to only retain XAI research papers by matching title, abstract, venue, and journal name of the article with the terms identified in the seminal survey on XAI by \citet{Adadi2018} (see \autoref{tab:XAI_words} in appendix for the full list).
This resulted in 3,533 articles.
Depending on the availability in S2ORC we either included the sentences of the abstract only, or sentences of the full article. 

We filtered the sentences to only keep those that included the target words, which resulted in  %
122,833 sentences. %
We also only kept sentences in which the target word was the subject. We used ScispaCy
~\cite{neumann-etal-2019-scispacy} for part-of-speech tagging and dependency parsing and extracted the subjects of the sentences (main and subordinate clauses). 
We then filtered the subjects for the target words, 
resulting in 12,893 sentences.
There are two reasons for this syntactic constraint:
First, subjects always have verbs, which are most indicative of mental state reference \cite{orrDevelopmentValidationMentalPhysical2022}.
Hence, we expect the most clear-cut mind attributions to be sentences in which the target word is the subject and the verb is a mental state reference (e.g. ``the model thinks''). 
Second, the syntactic structure of the sentence may have a strong effect on the clusters created in the semantic shift detection. %
In methods comparable to ours, sentences with a similar syntactic structure are sometimes grouped together independently of clear semantic patterns~\cite{laicherExplainingImprovingBERT2021}. 
As our main focus is on the semantic information reflected by the target word's context (specifically, the presence or absence of mind-attributing terms), we expected to achieve more semantic differentiation between clusters by restricting the syntactic variability in the sentences. 

To verify that we did not miss a very common mind-attributing usage of a target word that was not a subject, we also clustered %
all sentences. 
The clusters were similar in size to the ones obtained on the syntactically filtered corpus, indicating that there was not a widespread usage that we missed by restricting to subjects only. 
Moreover, the clusters with many sentences were more diverse, and did not consist of mind-attributing sentences only. 

\subsection{Extraction of Embeddings and Clustering}

We used SciBERT \cite{Beltagy2019SciBERT}, a BERT model trained on scientific texts, to produce contextualised word embeddings. 
BERT models use the transformer architecture and are trained on the masked language modeling and next sentence prediction tasks~\cite{devlinBERTPretrainingDeep2019}.
We used the \verb|scibert-scivocab-uncased| model from the HuggingFace implementation \cite{wolfHuggingFaceTransformersStateoftheart2020}. %
For each target word we used the sentence (main or subordinate clause) as input to SciBERT and extracted the resulting contextualised embeddings for each token in the sentence.
The input length for sentences is limited to 512 tokens for SciBERT, hence we only modelled sentences below this limit. %
Similar to \cite{mileticDetectingContactInducedSemantic2021}, we extracted the representations of the target words and averaged their values %
across the last four hidden layers to obtain a single contextualised embedding. 
If the target word was split into multiple subwords by SciBERT's tokeniser we averaged across subwords. 
We clustered the contextualised embeddings for each target word separately using affinity propagation \cite{AffProp}.
We opted for affinity propagation because this method 
performed better than k-means in a standard evaluation of semantic change detection \cite{martinc2020}. Moreover, we observed that the method tends to produce many small but coherent clusters.
This makes affinity propagation well-suited for studying fine-grained differences in target word usage.
We used the scikit-learn implementation with default parameters \cite{scikit-learn}.
We computed silhouette scores for each cluster assignment and excluded those with a negative score and those for which clustering did not converge. 
The silhouette score was $0.17 \pm 0.12$ %
(mean $\pm$ standard deviation). 
Previous work showed that even models with low silhouette scores %
represent semantic differences between clusters well \cite{martinc2020} and our qualitative analysis also confirms this. Hence, we decided not to tune the clustering parameters but use the default parameters as in previous work \cite{martinc2020, mileticDetectingContactInducedSemantic2021}.

\subsection{Selection of Clusters}
The clustering of the extracted embeddings resulted in 835 semantically different clusters of sentences.
There is no ground truth data to quantitatively evaluate the amount of mind-attributing language while reliably accounting for different meanings.
Hence, the assessment required manual qualitative analysis. 
Given the large number of clusters a full manual analysis was not feasible. 
We therefore excluded all clusters with less than 20 embeddings to avoid word usage that occurred in only a few papers. 
We also excluded all clusters with sentences from only one paper or one author.
These exclusion criteria resulted in 205 clusters. 
To further reduce the number we defined five additional criteria:
We included the 10 largest clusters to evaluate the most common usage of the target words in the corpus. 
Furthermore, we used two established quantitative metrics %
to identify clusters that were the most and the least mind-attributing. 
Specifically, we included the 10 clusters with the highest and lowest Mental-Physical Verb Norms (MPVN, \cite{orrDevelopmentValidationMentalPhysical2022}) score and the 10 clusters with the most and fewest matches with Mind Perception Dictionary (MPD, \cite{schweitzerLanguageWindowMind2021}) normalised by number of embeddings in each cluster.
This resulted in a final set of 46 clusters that we manually examined for linguistic content.

\subsection{Linguistic Analysis of Clusters}
\begin{figure}
    \centering
    \includegraphics[width=0.8\textwidth]{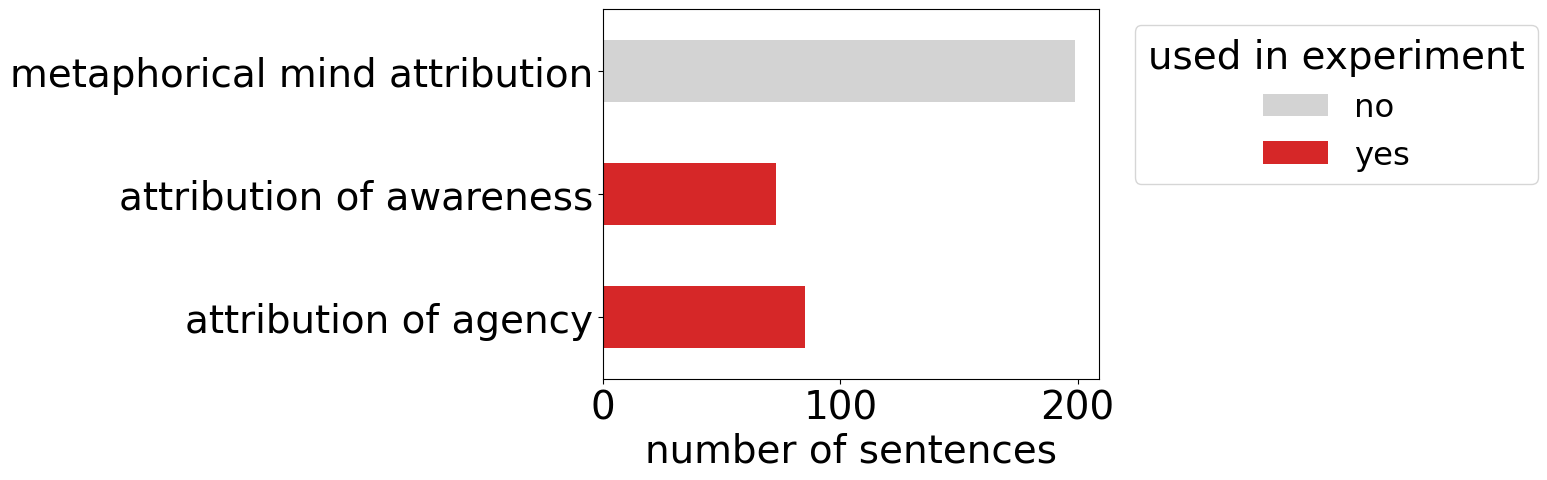}
   \caption{Number of sentences for three types of mind attribution identified in qualitative analysis. The red highlights indicates the terms that were used in the \textit{XAI mind attribution} condition of the experiment.}
   \Description{The figure shows a horizontal bar chart of three bars, one for each type of mind-attributing language. The size of the bars shows the number of sentences for each type. There are 199 sentences of metaphorical mind attribution in the corpus, 85 sentences that attribute agency to AI systems, and 73 sentences that attribute awareness. The bar for attribution of awareness and attribution of agency are highlighted in red because sentences from these are used in the experiment.}
   \label{fig:clusters}
\end{figure}

Our linguistic analysis of these clusters was based on the most distinctive keywords, the five most central sentences and, if needed, an exploration of all sentences contained in the cluster. 
We first selected the most distinctive keywords in each cluster by computing the term frequency--inverse document frequency (tf-idf) score for all words and bigrams as proposed by \citet{montariolScalableInterpretableSemantic2021}.
The central sentences were those represented by the five embeddings closest to the cluster centroids, similar to %
\citet{giulianelliAnalysingLexicalSemantic2020}. 
We provide the keywords and the central sentences of all manually analysed clusters in \autoref{app:clusteroverview}. 
By exploring the keywords and the central sentences of each cluster, we classified the shared commonality of the sentences. 

We first divided the clusters 
into one group of clusters that shared nouns, adjectives, determiners or adverbs, and a second group of clusters that shared verbs. The former group of clusters was mainly characterised by the direct neighbours of the target word independent of the semantic context. 
This suggests that the position and immediate context of a target word has a big influence on the clustering of SciBERT embeddings. 
As explained in \autoref{sec:corpus}, we expected the most clear-cut mind attributions in verbs and indeed the clusters that did not share a verb did not capture mind-attributing language. 
For completeness, we explain their commonalities in \autoref{app:clusteroverview}. 
The clusters that shared verbs provided the usage types we were looking for: We found three types of clusters with clear-cut mind attribution.

For each cluster we then determined if the sentences dealt with an AI system (the target of our analysis) or a XAI method. %
In our corpus, ``algorithm'' and ``model'' can refer to both 
and we used the following heuristic to distinguish between them:
We assumed that most XAI papers presented a novel XAI method and showed how the method worked on an exemplary established AI system.
Therefore, we assumed that detailed descriptions deal with a XAI method rather than an AI system. Similarly, we assumed that the possessive pronoun ``our'' indicated a novel XAI method.

In the following, we provide a detailed characterisation of the three types of mind-attributing language (for a quantitative summary, see \autoref{fig:clusters}). 

\subsubsection{Metaphorical Mind Attribution}

In XAI research, established mental state references are used as a metaphor for computational processes as shown in the following examples:
\begin{enumerate}[resume]
    \item \label{ex:learn-1} that the model has \textit{learned} to distinguish (...)
    \item \label{ex:predict} how often the model \textit{makes predictions}
\end{enumerate}
``To learn'' was the most prevalent example of metaphorical mind attribution in the corpus, found in several clusters.
Typical usage is shown in example (\ref{ex:learn-1}), from a coherent cluster of short subordinate clauses. 
Across all clusters, ``to learn'' is used metaphorically. It describes a computational process typical of AI systems, i.e. improving performance by continuously incorporating new data. This meaning is established in AI research and is clearly distinct from the one generally used for humans (to gain knowledge by study \cite{learndef}).
The same holds for the verb ``to predict'' and the construction ``make predictions'', which were keywords in four clusters.
Example (\ref{ex:predict}) shows the metaphorical usage typical of AI systems, i.e. computing an output given some input.
It differs from the general meanings of declaring in advance or thinking that something will happen \cite{predictdef}.
Some of the clusters of metaphorical mind attribution also included sentences 
with more technical descriptions. 
Consider the following two pairs of examples taken from one cluster each:
\begin{enumerate}[resume]
    \item \label{ex:learn-4} Then the classifier \textit{learns} to recognize them as members of the original class.
    \item \label{ex:learn-5} The classifier \textit{takes} a spectrum \textit{as input} and will \textit{output} a probability vector of belonging to one particular material.
\end{enumerate}
\begin{enumerate}[resume]
    \item \label{ex:predict-2} why the model \textit{predicted} the particular label for a single instance and what features were most influential for that particular instance
    \item \label{ex:predict-3} why the model \textit{processes} an unseen image as belonging to a specific domain and not the other
\end{enumerate}
In each pair of examples, the respective target word appears in similar contexts: the notion of membership expressed as ``members of'' in (\ref{ex:learn-4}) and ``belonging to'' in (\ref{ex:learn-5}), as well as the interrogative adverb ``why'' and the conjunction ``and'' in (\ref{ex:predict-2}) and (\ref{ex:predict-3}). 
This shows that the verbs ``to predict'' and ``to learn'' 
are used similarly to other technical verbs like ``to output'' and ``to process''. %
This reflects their technical meaning specific to AI systems.

\subsubsection{Attribution of Awareness}
The second type of usage contrasts the previous metaphorical mind attributions as there is no established definition of what it means for an AI system ``to consider'', ``to take into account'', or ``to focus''. 
 These verbs were used in examples such as the following:
\begin{enumerate}[resume]
    \item \label{ex:consider} The model \textit{considers} two DCF techniques (...)
    \item \label{ex:consider-2} The model \textit{takes into account} complex tasks (...)
    \item \label{ex:consider-3} The model \textit{focuses} on the vehicular ad-hoc part of the VBB application (...)
\end{enumerate}
Sentences such as (\ref{ex:consider}--\ref{ex:consider-3}) describe that the model was influenced by something by attributing awareness 
and were part of the same large cluster.  %
We also found the term ``to consider'' in a cluster of sentences that used ``our model'', indicating reference to XAI methods rather than AI systems.
The use of ``to consider'' both with AI systems and XAI methods suggests that it is a low-level mind attribution not restricted to complex methods. 

\subsubsection{Attribution of Agency}
The third type of usage was found in two clusters that attributed decision-making capabilities to the AI system.
A representative sentence is:
\begin{enumerate}[resume]
    \item \label{ex:decide} why the model \textit{made} a specific \textit{decision} for an instance
\end{enumerate}
Making a decision and acting on it is associated with the mind perception dimension of agency. 
We found a similar cluster for the target word ``algorithm'' and 
both had the highest number of conjunctions ``why'' and ``because'', as in example (\ref{ex:decide}), indicating that these sentences are about the AI system to be explained rather than a XAI method.

\subsubsection{Other Usage Types}
\label{sec:nomindverbs}
We found more diverse mind attributions to AI systems in a cluster %
mainly containing sentences about models making errors 
as shown below:
\begin{enumerate}[resume]
    \item \label{ex:mistake-1} when the model \textit{makes a mistake}
    \item \label{ex:mistake-2} that the model \textit{is forgetting} what it needs to look
    \item \label{ex:mistake-3} that the model \textit{knows} what it does not know

\end{enumerate}

We also analysed the non-mind-attributing verbs and identified two main themes. 
First, we found the usage of the non-mind-attributing verb ``to work'' in sentences that dealt with AI systems, exemplified below:
\begin{enumerate}[resume]
        \item \label{ex:work} how an algorithm \textit{works}
\end{enumerate}
This usage type was typical of two clusters in which 
the syntactic structure confirmed that the sentence was an explanation about an AI system.
We found a second theme of 
clusters that contained sentences providing a technical description of a method, which used non-mind-attributing verbs like ``to run'', ``to start'', and ``to terminate''. We consequently assumed that the clusters in this theme were mainly about XAI methods, and hence not the target phenomenon. Finally, six more clusters contained a non-mind-attributing verb in the keywords but did not show a clear pattern of how that verb was shared across the central sentences.

%% file: effect.tex
\section{Effect of Description of AI Systems on Perception of AI Systems}
To evaluate the impact of the mind-attributing language found in XAI research on the perception of AI systems [$H_I(1)$], we extended an established experiment.

\begin{figure}
    \centering
    \includegraphics[width=\textwidth,trim={0cm 11.5cm 0cm 3cm},clip]{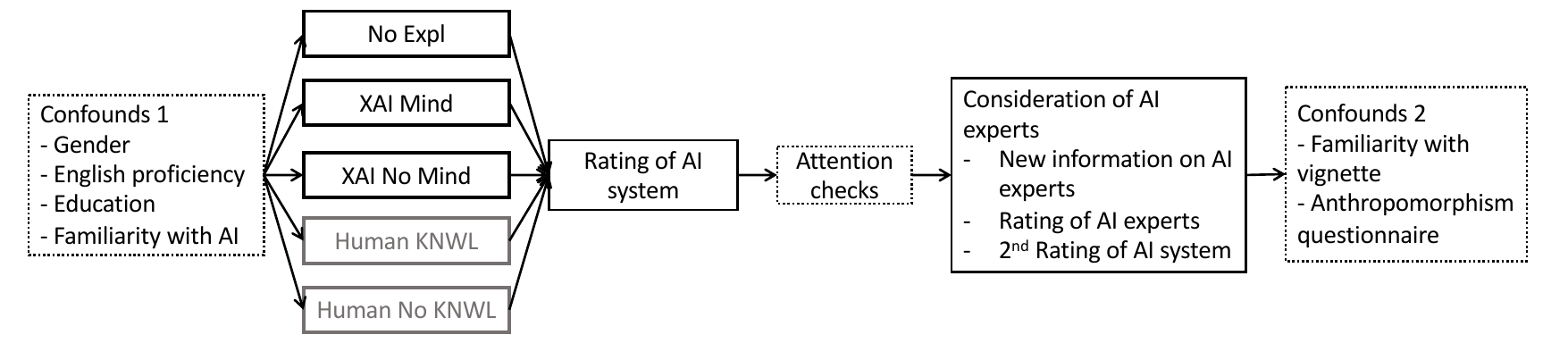}
    \caption{Overview of the experiment. 
    Branches indicate the five conditions in which the participants read different versions of the vignette. Grey boxes are the two reproduced vignette versions from \cite{stuartGuiltyArtificialMinds2021}, black boxes are the new versions based on language found in XAI research.}
    \label{fig:experiment}
\end{figure}

We were particularly interested in whether attribution of agency and awareness results in a shift of responsibility, i.e. that the AI system is held more  [$H_I(2)$] and the AI experts are held less responsible in the case of a negative outcome  [$H_I(3)$].
To investigate these questions, 
similar to prior work \cite{stuartGuiltyArtificialMinds2021}, we conducted a vignette-based experiment as shown in \autoref{fig:experiment}.

\subsection{Participants}
We recruited 303 participants on Amazon Mechanical Turk (AMT) and by reaching out to personal contacts, students, and staff at the local university via mailing lists and forums. On AMT, we restricted to the English-speaking countries Australia, Canada, New Zealand, the United States, and the United Kingdom to ensure comparable understanding of the English vignette.
Participants who failed two attention checks, responded with the same answer across all questions, or who did not respond to all dependent variable questions were excluded from further analysis.
199 participants remained after this step. We chose the sample size to be equivalent with the prior work \cite{stuartGuiltyArtificialMinds2021} to be able to reproduce their findings. 101 (51\%)  participants were women, 94 (47\%)  were men, two (1\%) chose other as their gender, and one (0.5\%) preferred not to say. The participants had a high self-rated proficiency of English with 149 (75\%)  of them being native speakers and 33 (17\%) reporting to be fluent. 133 (67\%) participants had completed a Bachelor's or higher degree.
The experiment design was approved by the institutional ethics review body.
Participants recruited via Amazon Mechanical Turk were remunerated USD\$1.50 for the, roughly, 10 minute survey; all other participants donated their time voluntarily.
Participants were given detailed information about the purpose and content of the study and gave their informed consent.
We stressed that they would read about a fictional event in which an AI system caused harm to people before starting the survey and that they could decide not to engage in the study.

\subsection{Experiment Design}
The experiment consisted of five conditions and two parts. Participants were randomly assigned to one of the five conditions and had to complete both parts. In the first part, participants were asked to rate the AI system, while in the second part they were asked to consider the AI experts, and subsequently reassess the AI system. 
There were two attention checks between the first and the second part in order to address the limited data quality in AMT \cite{peerDataQualityPlatforms2022}.
In addition, we included two sets of questions on potential confounding variables (one at the beginning and one at the end).

\subsubsection{Conditions}
The manipulation between the conditions consisted in different versions of a vignette 
proposed by \citet{stuartGuiltyArtificialMinds2021} that presents a situation in which an AI system causes harm to people.
We created three versions of the vignette with different depictions of the AI system: a baseline version without an explanation (\textit{no explanation}), a version with the mind-attributing explanation we found in XAI research (\textit{XAI Mind}), a version with an explanation avoiding mind attribution (\textit{XAI No Mind}).
Moreover, we included two reproductions of the original vignette.
The baseline \textit{no explanation} vignette is:
\begin{quote}
    \textit{Shill \& Co. is a farming company, which produces vegetables and fruits. The potato fields are managed by an artificial intelligence (AI) model. This year, the model uses a new fertiliser to increase the yield. The fertiliser has detrimental side-effects: it pollutes the groundwater in the area. Unfortunately, it is a very dry season. 
    The fertiliser does not get diluted by the rain and severely pollutes the groundwater. Many people in the area suffer serious health consequences.} 
\end{quote}

The \textit{XAI Mind} explanation contained the agency and awareness attributing expressions that we identified in our analysis of XAI research papers, namely ``to make decisions'' and ``to consider''. It reads as follows (emphasis in bold was not visible to participants, but is added to distinguish the conditions):  
\begin{quote}
    \noindent
    \textit{[...] The potato fields are managed by an artificial intelligence (AI) model, \textbf{which can make decisions}. [...] The fertiliser [...] pollutes the groundwater in the area. \textbf{The model considers this}. [...]}
    
\end{quote} 
For the \textit{XAI No Mind} version, we altered the sentences so as to create a coherent and plausible vignette with approximately the same amount of information, while avoiding mind attributions.
The resulting vignette reads the following:
\begin{quote}
    \textit{[...] The potato fields are managed by \textit{an artificial intelligence (AI) model}, \textbf{which takes agricultural data as input and performs calculations to find
    the fertilizer that maximizes yield}. This year, the model uses a new fertilizer which has detrimental side-effects: it pollutes the groundwater in the area. \textbf{This information has an influence on the model's output.} [...]}
\end{quote}
The sentences were adapted based on our corpus analysis through which we found multiple sentences that used ``to take as input'' to describe an AI system (e.g. example \ref{ex:learn-5}). 
We also identified the predicative use of ``influential'' in example \ref{ex:predict-2} as an alternative for the mind-attributing verb ``to consider''. 

We included two versions that were directly taken from the original experiment by \citet{stuartGuiltyArtificialMinds2021} to see whether we could reproduce their findings.
We call these conditions \textit{human-like knowledge} and \textit{human-like no knowledge}:
\begin{quote}
    \textit{[...] The potato fields are managed by \textit{Jarvis, a robot equipped with artificial intelligence}, \textbf{which can make its own decisions}. This year, Jarvis uses a new fertilizer to increase the yield. The fertilizer has detrimental side-effects: it pollutes the groundwater in the area. \textbf{Jarvis (knows / does not know) this}. [...]} 
\end{quote}

The full text of all vignette versions is in \autoref{app:vign_versions}. %

\subsubsection{Rating of AI System}
After reading the vignette, we asked participants to rate wrongness, responsibility, and awareness of the AI system (``Jarvis'' or ``the AI model'' depending on the condition):
The question ``How wrong was the action of [the AI system]?'' measured whether participants deemed the action to be wrong.
We included this question and show the analysis in \autoref{sec:wrongness} to be able to compare to previous work \cite{stuartGuiltyArtificialMinds2021} which has shown that wrongness depends more on the outcome than other moral judgements. In our further analysis, we will focus on responsibility and awareness .
In contrast to previous work \cite{stuartGuiltyArtificialMinds2021} we did not ask for blameworthiness but instead chose responsibility.
Responsibility is broader than blameworthiness and has a forward- and backward-looking notion \cite{limaHumanPerceptionsMoral2021, malleTheoryBlame2014}. 
We asked for responsibility directly with the question ``How responsible, if at all, is [the AI system] for the outcome?''[$H_I$(2)]. 
We finally asked for the perception of awareness with ``Was [the AI system] aware of the potential outcome?''[$H_I$(1)]. The term ``consider'' we used in the \textit{XAI Mind} attribution was less strong than ``know'' which was used in the original versions. 
Hence, we also asked for awareness instead of knowledge which was used in the original study to be able to capture more fine-grained differences. 
The response choices for each of these aspects ranged from not at all (1) to completely (7).

\subsubsection{Consideration of AI Experts}
In order to understand whether mind attributing language conceals the responsibility of the involved AI experts, 
we asked participants to consider the following information [$H_I$(3)].
\begin{quote}
     \textit{The AI experts who created [the AI system] \textbf{know} about the detrimental side effects of the fertiliser. } 
\end{quote}
We decided for this scenario because we wanted to see whether the mind-attributing explanation has an concealing effect even when the awareness of the AI experts is made very explicit.
As discussed by \citet{Lima2022}, AI experts might even negate their own involvement and thereby intentionally use AI systems as scapegoats.
In this study, however, we were interested in the effect that the mind-attributing explanations have on their own, i.e., even when the experts' involvement is clear. 
Participants were asked to rate wrongness, responsibility and awareness of the AI experts in the same way as above. 
Afterwards, we asked the participants to reassess the AI system's wrongness, responsibility and awareness, to see whether considering the AI experts changed their judgements [$H_S$].

\subsubsection{Attention Checks}
Amazon Mechanical Turk is widely used in academic research, despite the fact that responses obtained using this platform can have limited data quality with respect to attention, comprehension and dishonesty \cite{peerDataQualityPlatforms2022}.
To increase attention and honesty we extended the sample to volunteer participants with a direct or close personal contact.
Moreover, participants completed attention checks in which we asked comprehension questions about the vignette. 
Participants who were unable to give the correct answer were excluded from the analysis but were still allowed to continue the survey and were remunerated.
We only included responses from 199 out of 304 participants who passed these checks.
Participants' comments like ``I'm interested in knowing about lawsuits and their outcomes for similar situations. Thank you.'' and ``interesting and thought provoking survey'' suggest that this filtering was effective and that participants were indeed engaged in the survey and motivated to provide meaningful answers. 

\subsubsection{Potential Confounds}
The first set of questions on confounding variables were about participants' gender, proficiency in English, highest level of education, and familiarity with AI. 
At the end of the experiment, we asked participants whether they were familiar with the vignette and 
used the IDAQ questionnaire in order to measure the tendency to anthropomorphise \cite{waytzWhoSeesHuman2010}.
Lastly, participants could provide a free text comment.

\subsection{Results}

We used ordinal regression models to fit the ratings of awareness and responsibility following the tutorial by \citet{burknerOrdinalRegressionModels2019}. 
In contrast to statistical tests like ANOVA used in previous research \cite{stuartGuiltyArtificialMinds2021, shankAttributionsMoralityMind2018}, these models do not assume the data to be metric which ordinal Likert ratings are not. Using ANOVAs despite the false assumption can lead to missing the correct detection of differences, wrong effect-size estimates, bigger Type 1 error rates, and in the worst case the reversal of differences \cite{liddellAnalyzingOrdinalData2018}.
Instead, we used cumulative ordinal regression models that assume an underlying latent continuous variable that is partitioned into the seven observed rating values. The six thresholds separating the ratings were estimated from the data. We fit the ratings collected in our experiment to a total of five models explained below.

The models estimate the causal influence between the vignette versions and the ratings of awareness, responsibility and wrongness. Any influence of confounding variables (including observed confounds like education and AI familiarity, and unobserved variables like cultural or moral values) on the ratings was ruled out as participants were randomly assigned to the five vignette versions. We verified that the observed variables are similarly distributed across the conditions by plotting the distributions and visually inspecting them (see \autoref{app:confounds} for plots).

\subsubsection{Reproduction of Findings by \citet{stuartGuiltyArtificialMinds2021}}
We recreated two vignette versions as they were used in a previous study \cite{stuartGuiltyArtificialMinds2021} in order to verify whether we could reproduce their findings with new participants, a different analysis (ordinal regression model instead of ANOVA), and adapted questions (responsibility instead of blame, and awareness instead of knowledge).
\begin{figure}
    \begin{subfigure}[t]{0.32\textwidth}
         \centering
         \includegraphics[width=\textwidth]{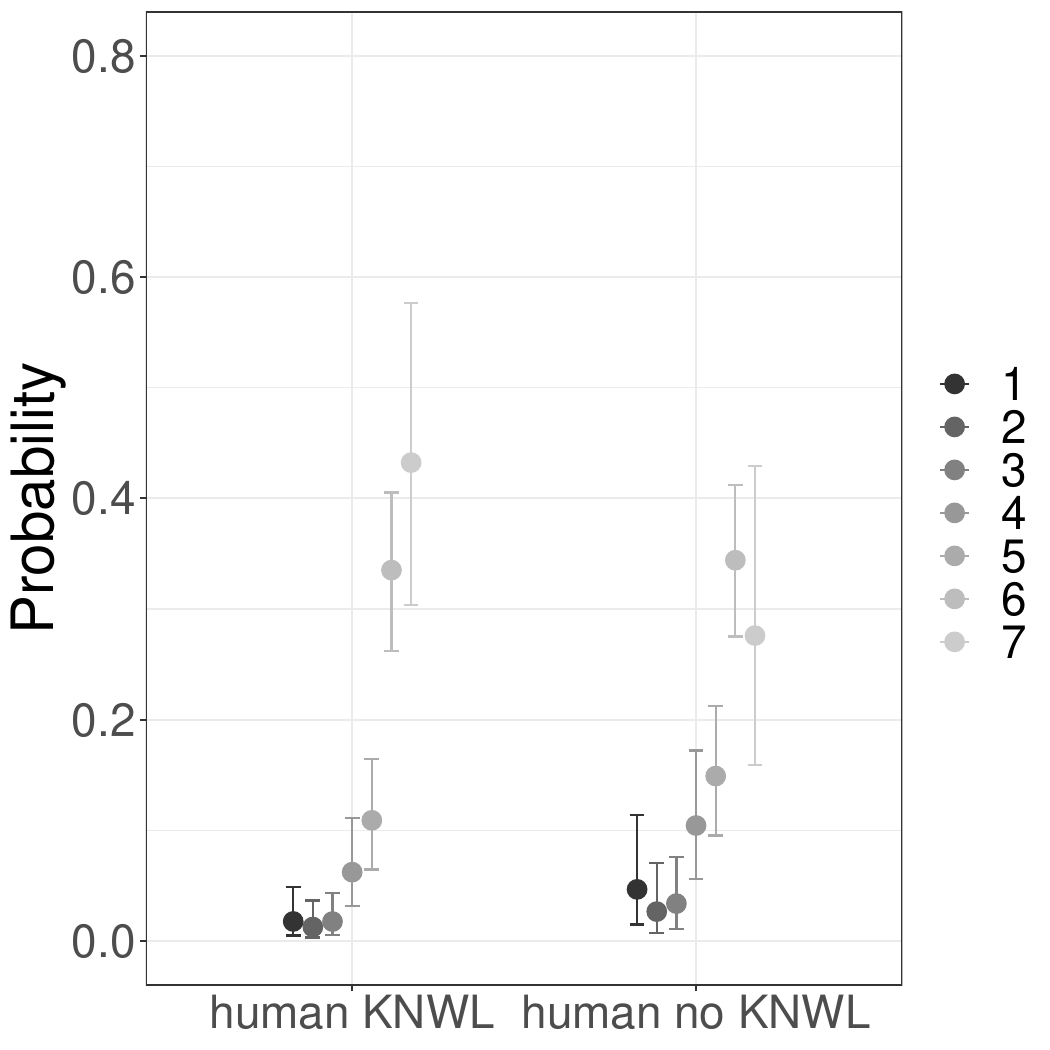}
         \caption{Wrongness}
         \label{fig:AIwrong1_Reproduction}
    \end{subfigure}
    \begin{subfigure}[t]{0.32\textwidth}
         \centering
         \includegraphics[width=\textwidth]{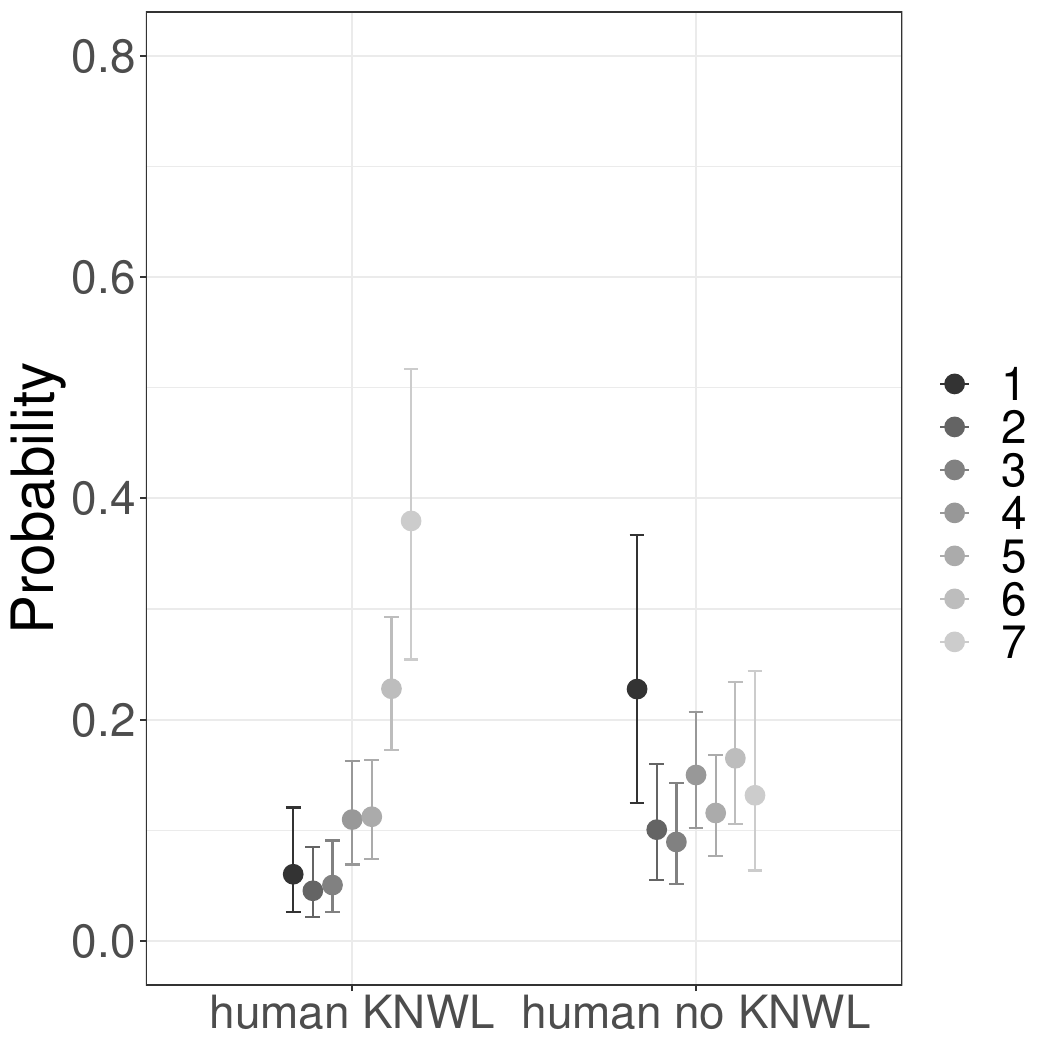}
         \caption{Responsibility}
         \label{fig:AIresp1_Reproduction}
    \end{subfigure}
    \begin{subfigure}[t]{0.32\textwidth}
         \centering
         \includegraphics[width=\textwidth]{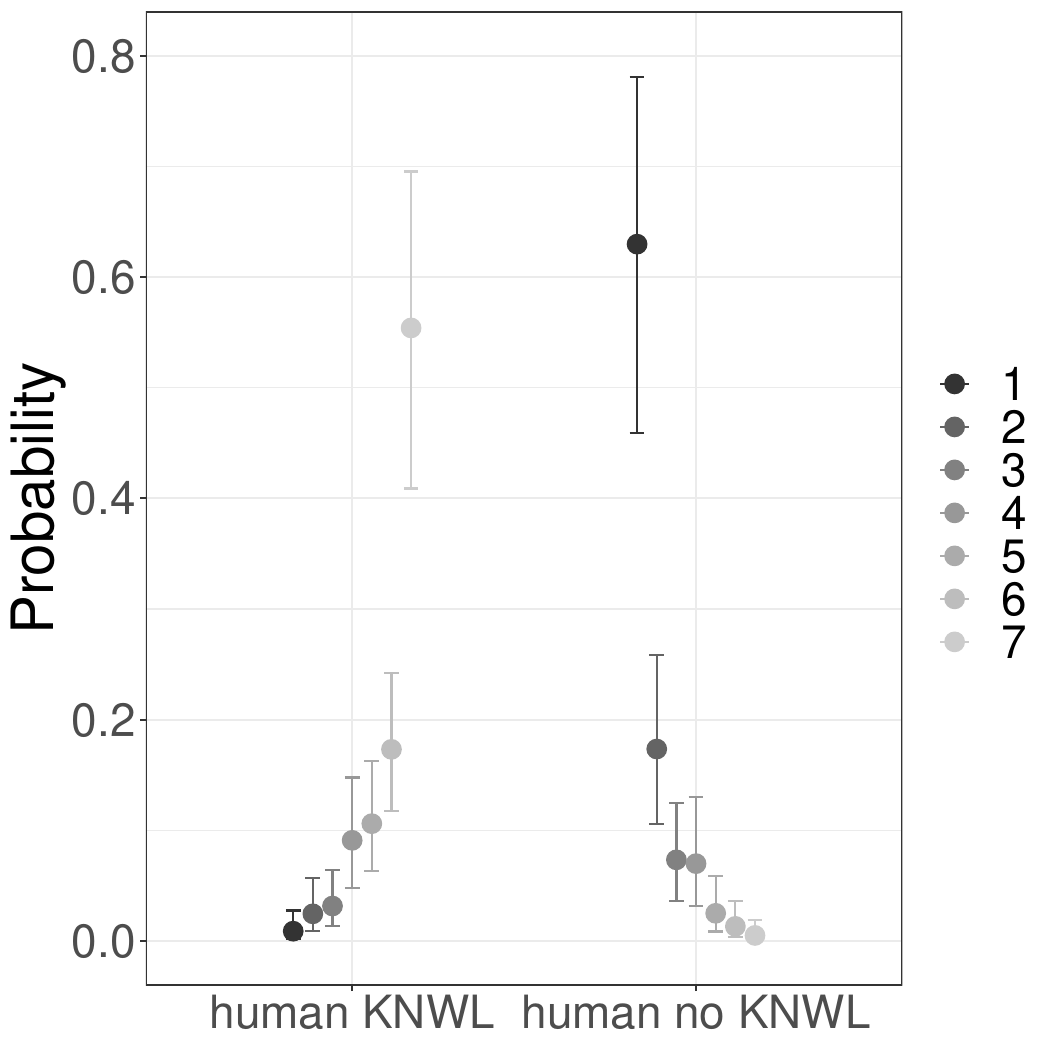}
         \caption{Awareness} 
         \label{fig:AIaware1_Reproduction}
    \end{subfigure}
    \caption{Marginal effect of reproductions on first ratings of AI system. The reproductions are taken from \citet{stuartGuiltyArtificialMinds2021} and depict a human-like robot that either ``knows'' (\textit{human KNWL}) or ``does not know'' (\textit{human no KNWL}) about the detrimental side effects. Participants rate wrongness, responsibility, and awareness on a 7-item Likert scale for which the probabilities are given. }
    \label{fig:Reproduction}
\end{figure}

\begin{table}[h!]
\centering
\begin{tabular}{l c c c} 
 \hline
  & \textbf{Estimate (Error)} & \textbf{l-95\% CI} & \textbf{u-95\% CI}  \\ 
 \hline
 Wrongness & -0.43 (0.27)&   -0.96 &  0.10 \\ 
 Responsibility & -0.81 (0.26)  &  -1.34  &  -0.27 \\
 Awareness & -2.70 (0.31)&    -3.32  &  -2.12 \\
 \hline
\end{tabular}
\caption{
Estimates, lower bounds and higher bounds of 95\% credible intervals for coefficients in the ordinal regression models that indicate the difference between the two reproductions. The \textit{human-like knowledge} condition is the reference, hence the coefficient indicates whether the ratings of participants in \textit{human-like no knowledge} differ. The coefficient estimates are taken from three regression models, one for each rating (wrongness, responsibility, awareness).}
\label{table:reproduction}
\end{table}

We fitted three regression models, one for the first ratings of each wrongness, responsibility and awareness of the AI system. We used dummy coding with the baseline \textit{human-like knowledge} as a reference. We show the marginal effect of the two reproduced vignette versions in \autoref{fig:Reproduction}, and the coefficient estimates for the difference in \autoref{table:reproduction}. 
\autoref{fig:AIwrong1_Reproduction} shows the estimated probabilities for the seven rating levels for wrongness. There was a slight difference between the \textit{human-like knowledge} condition and the \textit{human-like no knowledge} condition, as participants in the latter were less likely to respond with the highest rating (7). This is also reflected in the slight difference in the point estimate shown in the first row of \autoref{table:reproduction}.
The effect of the original vignette versions is even more distinct for the ratings of responsibility shown in \autoref{fig:AIresp1_Reproduction}.
The participants in the \textit{human-like no knowledge} condition were most likely to rate the responsibility of the AI system with the highest value (7), while participants in the \textit{human-like no knowledge} condition were most likely to rate the responsibility with the lowest value (1). This is also shown by the clearly negative 95\% credible interval of the estimate of the difference between the two reproductions in the second row in \autoref{table:reproduction}. 
Similarly, participants in the \textit{human-like no knowledge} condition were most likely to rate the system's awareness with the lowest value (1) compared to participants in the \textit{human-like knowledge} condition who were most likely to rate awareness with the highest value (7) (\autoref{fig:AIaware1_Reproduction}). On the latent scale, their ratings were 2.70 SD lower with the 95\% CI between -3.32 and -2.09 (shown in the last row of \autoref{table:reproduction}).
In the original study \cite{stuartGuiltyArtificialMinds2021}, 
the authors report a significant main effect of knowledge on wrongness, blame and knowledge. Our data reproduces the findings for responsibility (matched with blame) and awareness (matched with knowledge) as well as, somewhat less clearly, for wrongness. %

\subsubsection{Awareness of AI System}
\begin{figure}
    \centering
    \includegraphics[width=0.5\textwidth]{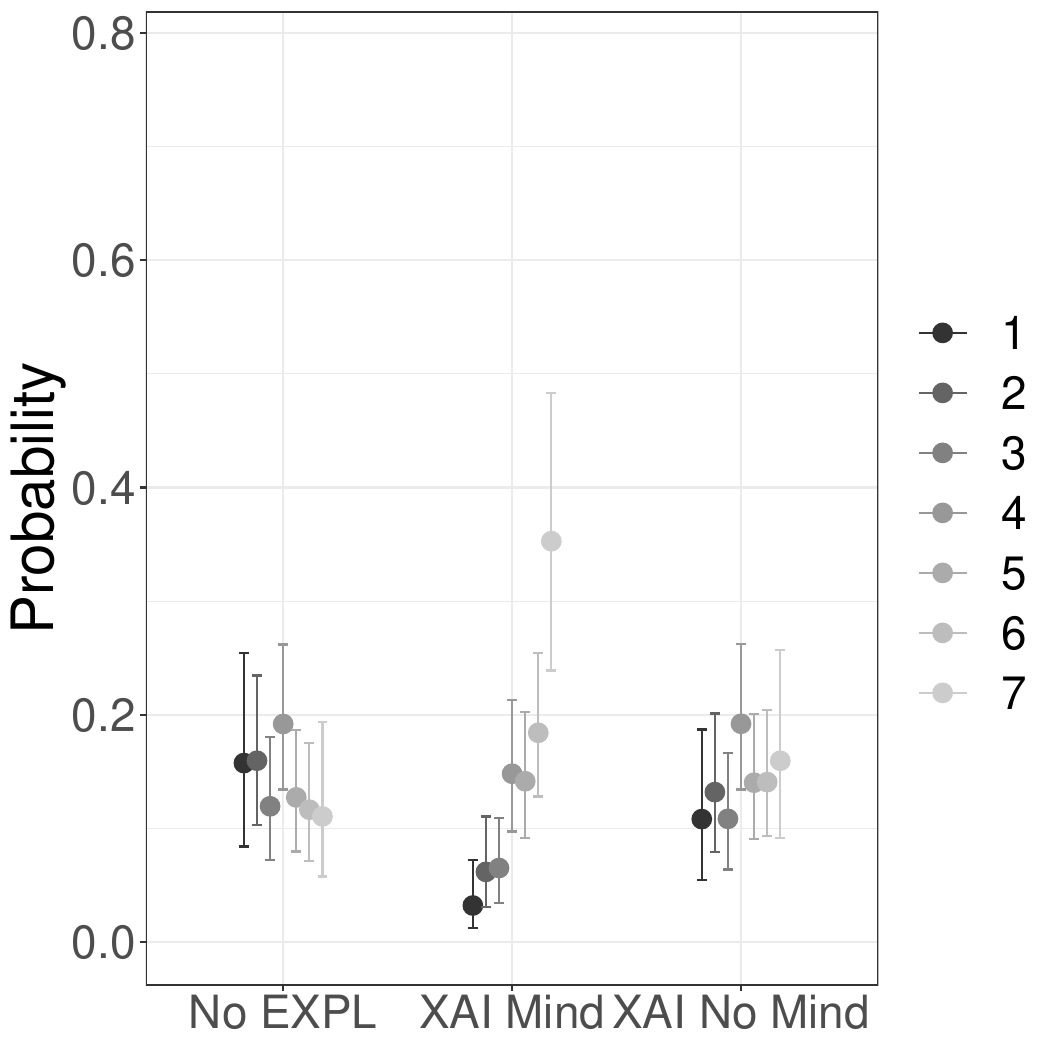}
    \caption{Marginal effects of vignette version (\textit{no explanation}, \textit{XAI Mind}, and \textit{XAI No Mind}) on first ratings of AI awareness.}
    \label{fig:AIaware1}
\end{figure}
\begin{table}[h!]
\centering
\begin{tabular}{l c c c} 
 \hline
 Awareness & Estimate (Error) & l-95\% CI & u-95\% CI  \\ 
 \hline
 XAI Mind & 0.84 (0.23) &    0.41  &   1.28 \\ 
 XAI No Mind & 0.23 (0.22) &    -0.20   &  0.67 \\
 \hline
\end{tabular}
\caption{Estimates, lower bounds and higher bounds of 95\% credible intervals for coefficients for the ordinal regression model of awareness ratings. The first row indicates the difference between the \textit{XAI mind} condition and the \textit{no explanation} baseline. Second row indicates the difference between the \textit{XAI no mind} condition and the \textit{no explanation} baseline.}
\label{table:AwarenessCoeff}
\end{table}
We fitted an ordinal regression model for the first rating of awareness of the AI system with the vignette version as the only predictor to investigate $H_I(1)$. The marginal effect of the vignette version is shown in \autoref{fig:AIaware1}. Participants in the \textit{XAI Mind} condition were more likely to rate the awareness of the AI system with the highest value (7) than participants  in the \textit{no explanation} condition. This is also reflected in the point estimate. On the latent continuous rating scale, their ratings were 0.84 SD higher compared with participants who received no explanation (first row in \autoref{table:AwarenessCoeff}). 
The 95\% CI of this parameter was between  0.39 and 1.29, and did not include zero. We can therefore conclude with at least 95\% probability that participants who read the mind-attributing explanation attribute more awareness to the AI system than did participants who read the baseline vignette. Participants in the \textit{XAI No Mind} condition rated the AI system's awareness more similar to participants who received no explanation. The difference (0.23 SD higher on the latent awareness scale, shown in the second row in \autoref{table:AwarenessCoeff}) was smaller than for the \textit{XAI mind} condition. %

\subsubsection{Responsibility}
\begin{figure}
    \centering
    \includegraphics[width=0.9\textwidth]{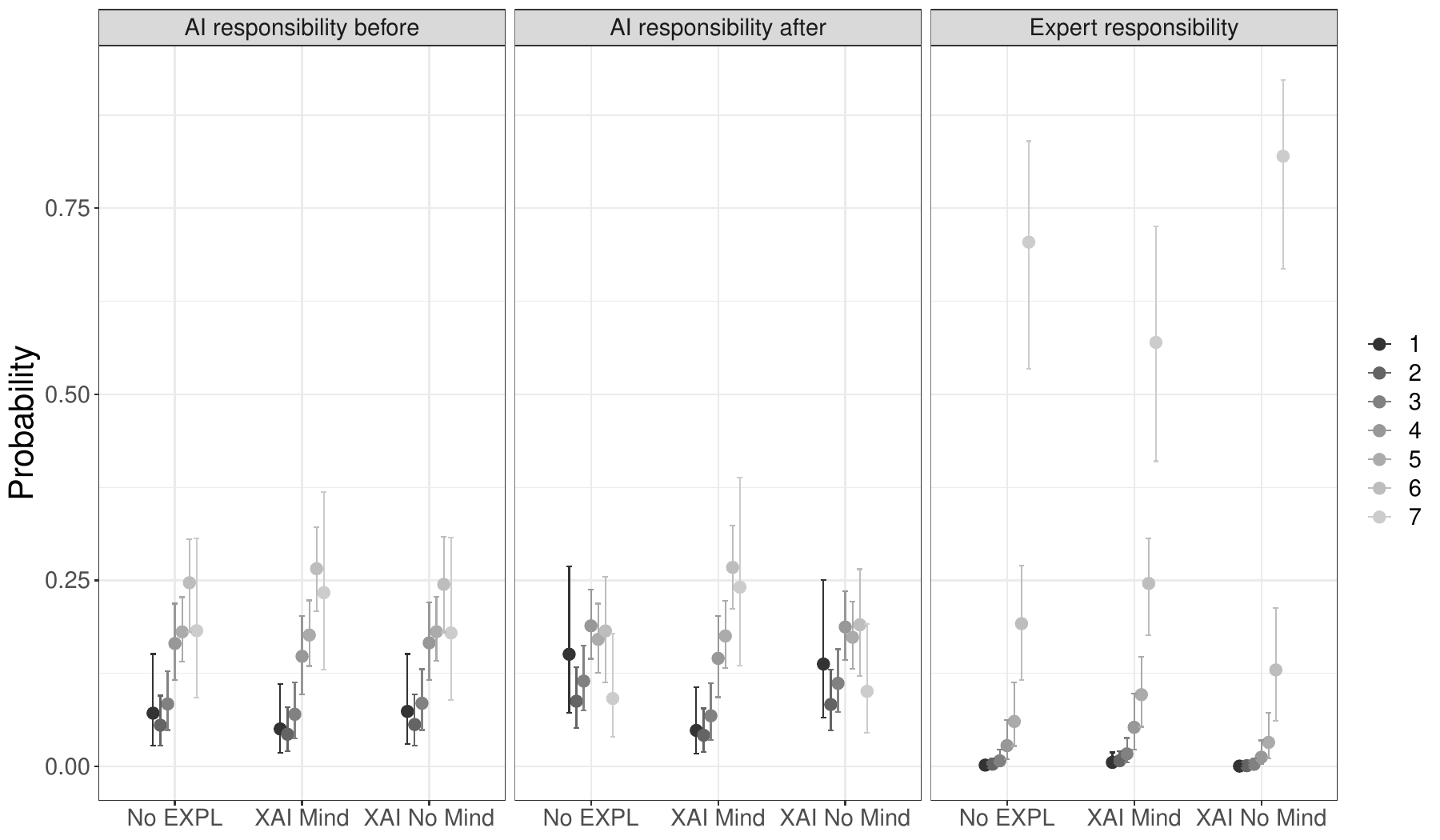}
    \caption{Marginal effects of vignette version(\textit{no explanation}, \textit{XAI Mind}, and \textit{XAI No Mind}) on ratings of responsibility before (left) and after (middle) considering the AI experts, as well as for rating the AI experts.}
    \label{fig:resp}
\end{figure}
\begin{table}[h!]
\centering
\begin{tabular}{l c c c} 
 \hline
 \textbf{Responsibility} & \textbf{Estimate (Error)} & \textbf{l-95\% CI} & \textbf{u-95\% CI}  \\ 
 \hline
 Before XAI Mind & 0.18 (0.28) &    -0.38  &   0.72 \\ 
 Before XAI No Mind & -0.01 (0.29) &    -0.58  &   0.55 \\
 Expert & 1.44 (0.26) &     0.94  &   1.95 \\
 Expert XAI Mind & -0.54 (0.34) &    -1.21  &   0.11 \\
 Expert XAI No Mind & 0.40 (0.38) &    -0.34  &   1.13 \\
 After & -0.43 (0.23) &    -0.87  &   0.02 \\
 After XAI Mind & 0.46 (0.32) &   -0.17  &   1.08 \\
 After XAI No Mind & 0.07 (0.32) &    -0.57  &   0.69 \\
 \hline
\end{tabular}
\caption{Estimates, lower bounds and higher bounds of 95\% credible intervals for coefficients for the ordinal regression model of responsibility ratings. The rows show the estimate, lower bounds and higher bounds of 95\% credible intervals for the coefficients that code the comparison to the baseline which is the first rating of responsibility \textit{before} considering the experts for the \textit{no explanation} vignette version.}
\label{table:ResponsibilityCoeff}
\end{table}
We also fitted an ordinal regression model for the three ratings of responsibility (ratings before and after considering the experts, and rating of expert responsibility). We again used dummy coding for the vignette version, and included an additional predictor for the three ratings. 
While ratings were similar for the three vignette version at first ([$H_I$(2)], left panel in \autoref{fig:resp}), they were more different when participants rated the AI experts ([$H_I$(3)], right panel in \autoref{fig:resp}), and subsequently reassessed the AI system ([$H_S$], middle panel in \autoref{fig:resp}). 
The coefficient estimates for the first rating show that 
participants in the \textit{XAI Mind} condition first rated the AI system's responsibility slightly higher (0.18 SD on the latent continuous rating scale, first row of \autoref{table:ResponsibilityCoeff}) than participants in the \textit{no explanation} condition. This difference is slightly bigger than the comparison between participants in the \textit{XAI No Mind} condition and \textit{no explanation} condition (-0.02 SD on the latent continuous rating scale, second row of \autoref{table:ResponsibilityCoeff}).

Before we explain the results of the second rating of responsibility (middle panel in \autoref{fig:resp}), we consider the ratings of expert responsibility (right panel in \autoref{fig:resp}). In general, participants rated the responsibility of the experts higher than the AI system's (1.44 SD on the latent continuous scale, third row of \autoref{table:ResponsibilityCoeff}). %
However, participants in the \textit{XAI Mind} condition are less likely to rate the experts' responsibility with the highest value (7) than those in the \textit{XAI No Mind} and \textit{no explanation} condition. This is also shown in the coefficient estimates in the fourth and fifth row of \autoref{table:ResponsibilityCoeff}.

Subsequently, the difference in assessing the experts' responsibility was also reflected in the second rating of the AI system's responsibility (middle panel in \autoref{fig:resp}). Across conditions, participants reduced their responsibility ratings of the AI system after considering the experts. On the latent continuous scale, the second rating is -0.43 SD smaller than the first, and the 95\% CI is between -0.89 and 0.02 (sixth row in \autoref{table:ResponsibilityCoeff}). However, this general trend is not true for participants in the \textit{XAI Mind} condition. They instead are most likely to rate the responsibility with the second-highest value (6), resulting in an increased rating of 0.46 SD higher than those who received no explanation (seventh row in \autoref{table:ResponsibilityCoeff}). 
As in the first responsibility rating, the ratings of participants in the \textit{XAI No Mind} condition are similar to those who did not receive an explanation.

%% file: discussion.tex
\section{Discussion}

\subsection{Mind Attribution is Prevalent in XAI Research}
Our data-driven analysis revealed that XAI researchers use 
mind-attributing descriptions
([$H_P$], see \autoref{fig:clusters}).
This includes metaphorical mind attribution statements but also attributions of awareness and agency.
Previous work shows that the use of mind-attributing language meaningfully indicates differences in mind perception, as well as downstream consequences \cite{schweitzerLanguageWindowMind2021, orrDevelopmentValidationMentalPhysical2022}. 
Similarly, our finding show that the mind-attributing explanation increased the likelihood that participants will rate an AI system as aware ([$H_I$(1)], \autoref{fig:AIaware1}).
However, researchers might also use mental state references to explain the behaviour of their models without having to specify the detailed true underlying causes.
This usage of mind attribution is in line with the intentional stance \cite{Dennett1971} that helps to explain other agents by attributing beliefs and desires. 
For authors and readers of XAI research papers this non-literal meaning is likely clear from the context in which these statements were made:
More complete non-mind-attributing explanations about the true underlying processes might allow them to assess the AI system accordingly.
It will be interesting to see whether authors of XAI research papers perceive AI systems as mindful, aware agents.

\subsection{Differences in Terminology Might Explain Mixed Results in AI Responsibility Research}
Complementing earlier work \cite{langerLookItComputer2022}, we found that despite using the same terms (e.g. ``AI model'') the perception of the AI system's awareness depends
on giving an explanation and whether the explanation contains mind-attributing language.
\citet{shankAttributionsMoralityMind2018} have revealed that adding additional information on the algorithm had no effect on awareness ratings in their study. However, they found an effect of their seven tested scenarios in which the amount of mental state references used in the algorithm description differed. 
Our results suggest that the scenario's varying levels of mind attribution could explain the dependence of awareness ratings on the scenario.

Moreover, the difference in participants' ratings between the reproductions and our adaptations is noteworthy:
When the AI system is depicted as a named human-like robot, the ratings of awareness were higher than in the more realistic descriptions used in XAI literature (compare \autoref{fig:AIaware1_Reproduction} and \autoref{fig:AIaware1}). Similarly, responsibility ratings were higher for the human-like depiction than for all of our adaptations (compare \autoref{fig:AIresp1_Reproduction} and \autoref{fig:resp}). This suggests that the perception of AI systems in vignette-based studies depends on the depiction of the AI system. 
If the vignette describes a human-like AI system as in \citep{malleAISkyHow2019,malleWhichRobotAm2016,hongRacismResponsibilityAutonomy2019, stuartGuiltyArtificialMinds2021}, the systems are perceived in a human-like way.
In contrast, a more realistic description results in lower ratings of responsibility and awareness in our study, resonating with earlier work by \citet{shankAttributionsMoralityMind2018}.
This is an important finding for researchers investigating whether and how AI systems are held responsible using vignette-based experiments.
The terms used to describe the AI system in the vignette is likely to affect responsibility ratings.
In order to increase robustness and replicability, researchers should therefore always report the exact descriptions used, as a modified description might not reproduce previous findings (as our study shows). 

\subsection{Mind-Attributing Explanations Shift Responsibility}
Inspired by previous work \cite{Lima2022, shankAttributionsMoralityMind2018, robbinsMisdirectedPrincipleCatch2019}, we hypothesised that AI systems are held more responsible when an explanation is provided, and even more so when the explanation uses mental state references. 
We cannot 
support this hypothesis based on our experimental results ([$H_I(2)$], see \autoref{fig:resp}):
The responsibility of the AI system is likely driven by other factors which might include the outcome \cite{stuartGuiltyArtificialMinds2021}, intentionality and obligation \cite{malleTheoryBlame2014}. %
Furthermore, who is held responsible depends on the broader historical, political and social context \cite{delgrecoRevisitingAttributionTheory2021}.
While our results did not support the hypothesis directly, there was an indirect effect of explanations on the decrease of the AI system's responsibility rating
upon participants learning about the human contribution towards the harmful action ([$H_S$], see \autoref{fig:resp}). 
In general, participants reduced their responsibility rating for the AI system after considering the AI experts.
In contrast, those who had read the mind-attributing explanation gave slightly increased responsibility ratings (see middle panel in \autoref{fig:resp}, and seventh row of \autoref{table:ResponsibilityCoeff}).
Moreover, these participants also rated the responsibility of the experts lower than participants who received no explanation or read the non-mind-attributing explanations ([$H_I(3)$], see right panel in \autoref{fig:resp}, and fourth row of \autoref{table:ResponsibilityCoeff}).
This suggests that mind-attributing explanations as found in XAI research can conceal the responsibility of the humans who created the AI systems as suggested by \citet{Lima2022}. 

\subsection{The Necessity to Be Mindful About Mind Attribution}
Our results call for a cautious use of mind attribution in explanations about AI systems because it might impact how we interact with AI systems in at least two ways:\\
First, the language used in XAI research can have an effect beyond its scope as these descriptions ``trickle down'' to educational materials, media reports or regulators \cite{longWhatAILiteracy2020a, nataleImaginingThinkingMachine2020, kapoorHowReportBetter,gizinskiMAIRFrameworkMining2021}. 
Previous work has shown that the terms used for AI systems (such as ``artificial intelligence'' or ``statistical model') affect perception and evaluation of AI systems \cite{langerLookItComputer2022}. 
Our findings suggest that mind-attributing explanations should be used carefully in public-facing documents because they might unintentionally strengthen the perception of AI systems as mindful, aware agents. 
Other language contexts might counteract this effect: For example, sentences in which a human user or system designer is the subject and uses an AI system as a tool might strengthen the perception that humans, not AI systems, hold the mindful, responsible agency. 
As such, XAI researchers have the opportunity and power to shape the perception of AI systems when choosing the language to explain them.
Whether and which language usage in XAI research makes its way to the general public and how this changes the perception of AI systems, remains to be studied.

Second, mind-attributing explanations might be intentionally used by AI system designers to distance themselves from any negative outcome and use the AI system as a scapegoat as outlined by \citet{Lima2022}. Previous work has shown that misleading explanations successfully change the users' trust in an untrustworthy AI system \cite{lakkarajuHowFoolYou2020, banovicBeingTrustworthyNot2023a}.
In contrast, trustworthy explanations should redirect mind attributions away from the AI system and towards the involved humans. In our study, we 
redirected the mind attribution towards the AI experts by stating that they knew about the potential harm. 
This redirection serves two purposes: Firstly, it increases the degree to which AI experts are held responsible. In contrast to previous work, in which distal entities with respect to the harmful action receive a smaller share of responsibility than proximate entities \cite{shankAttributionsMoralityMind2018}, participants in our study rated the responsibility of the AI experts higher than that of the AI system (see \autoref{fig:resp}).
The second purpose of redirecting mind attribution towards the involved humans is to help answer the question of who amongst, potentially multiple, involved humans is responsible \cite{nyholmAttributingAgencyAutomated2018}. 
AI experts are only one group of humans involved in the application of AI systems. Other groups include the owners, users, and regulators of AI systems. 
Current XAI methods fall short in addressing these two explanatory needs 
but there are approaches to show the human involvement:
One exploration of so-called social transparency %
is provided by \citet{Ehsan2021} 
who extended XAI methods to include the socio-organisational environment of the AI system which helps to assess the responsibility of the users. Another very different approach are datasheets for datasets that help to understand the human decisions in data collection~\cite{Gebru2018}. 
Ultimately, our results call for the development of novel XAI methods that place an even stronger focus on the socio-technical perspective by surfacing who interacts with AI systems.

\section{Conclusion}
Explanations about AI systems that attribute a mind might lead to the AI system being solely held responsible, concealing the responsibility of the involved humans. 
We conducted a large-scale computational analysis of the language used in XAI research and found three kinds of mind attribution: 
metaphorical mind attribution (``to learn''), attribution of awareness (``to consider''), and attribution of agency (``to make decisions''). 
In a vignette-based experiment, we tested the impact of an explanation that contained awareness and agency attributing sentences and found increased ratings of awareness of the AI system.
Moreover, we found that redirecting mind attributions towards the involved humans increases their share of responsibility.
Interestingly, participants who had read the mind-attributing explanation drawn from XAI research
were reluctant to lower responsibility despite considering the involved humans.
The findings call for a cautious use of mind-attributing descriptions about AI systems and a stronger focus on the involved humans in explanations.

%% file: appendix.tex
\section{List of Target Words}
\autoref{tab:target_words} shows all target words selected from the 12 most highly cited AI research articles and verified by manual annotation. 
To ensure the broadest coverage, we collected main constituents of multiword terms %
(e.g. "network" instead of "neural network" and "convolutional neural network"), abbreviations (e.g. "CNN"), terms for specific architectures in their basic form (e.g. "Inception" and not "Inception-v4"), and general terms like "method", "tool", "algorithm". 
\begin{table}[H]
\begin{tabular}{p{3cm}p{5cm}p{2cm}p{2cm}p{2cm}}
\toprule
                     target word &                                    different forms &  count all sentences &  count subject sentences &  silhouette coefficient \\
\midrule
                       ``model'' &                                         ``models'' &                48811 &                     4975 &                    0.04 \\
                   ``algorithm'' &                                     ``algorithms'' &                27197 &                     4834 &                    0.06 \\
                     ``network'' &                                       ``networks'' &                16578 &                     1547 &                    0.10 \\
                  ``classifier'' &                                    ``classifiers'' &                 4162 &                      426 &                    0.09 \\
                ``architecture'' &                                  ``architectures'' &                 3598 &                      347 &                    0.08 \\
                       ``graph'' &                                         ``graphs'' &                 8175 &                      238 &                    0.13 \\
              ``classification'' &                                ``classifications'' &                 5581 &                      101 &                    0.05 \\
                     ``decoder'' &                                       ``decoders'' &                  560 &                       81 &                    0.12 \\
                         ``CNN'' &               ``CNNs'', ``ConvNet'', ``ConvNets'', ``convnet'' &                  617 &                       71 &                    0.09 \\
                    ``ensemble'' &                                      ``ensembles'' &                 1163 &                       66 &                    0.15 \\
                  ``regression'' &                                    ``regressions'' &                 2232 &                       55 &                    0.13 \\
                     ``encoder'' &                                       ``encoders'' &                  930 &                       48 &                    0.24 \\
                    ``pipeline'' &                                      ``pipelines'' &                  701 &                       47 &                    0.11 \\
                         ``RNN'' & ``recurrent neural network'', ``RNNs'', ``recurrent neural networks'' &                  241 &                       15 &                    0.15 \\
                      ``ResNet'' &                            ``ResNets'', ``WideResNet'' &                  143 &                        8 &                    0.27 \\
                         ``VAE'' & ``VAEs'', ``variational auto-encoder'', ``variational auto-encoders'' &                  120 &                        7 &                    0.50 \\
                         ``GAN'' & ``generative adversarial network'', ``GANs'', ``generative adversarial networks'' &                  173 &                        6 &                    0.16 \\
                         ``ICA'' &            ``iterative classification algorithm '' &                   66 &                        4 &                    0.37 \\
                 ``transformer'' &                                   ``transformers'' &                   90 &                        4 &                    0.14 \\
                   ``GoogLeNet'' &                                      ``GoogleNet'' &                   41 &                        3 &                    0.44 \\
                ``auto-encoder'' &       ``auto-encoders'', ``autoencoder'', ``autoencoders'' &                   61 &                        3 &                    0.26 \\
                         ``MLP'' & ``MLPs'', ``Multi-layer perceptron'', ``Multi-layer Multi-layer perceptrons'' &                  100 &                        3 &                    0.17 \\
                         ``GNN'' & ``GNNs'', ``graph neural network'', ``graph neural network'' &                   42 &                        2 &                     NaN \\
                         ``Net'' &                                           ``Nets'' &                 1392 &                        1 &                     NaN \\
                         ``GCN'' & ``graph convolutional network'', ``GCNs'', ``graph convolutional networks'' &                   31 &                        1 &                     NaN \\
    ``artificial intelligence'' &                                         ``AI'' &                    0 &                        0 &                     NaN \\
``Restricted Boltzmann Machine'' &             ``Restricted Boltzmann Machines'', ``RBM'' &                    0 &                        0 &                     NaN \\
                       ``DCGAN'' &                                         ``DCGANs'' &                    1 &                        0 &                     NaN \\
                        ``WGAN'' &                                             &                   27 &                        0 &                     NaN \\
\bottomrule
\end{tabular}
\caption{Full list of target words. The different forms were included to achieve broad coverage. The third column is the number of sentences with the target word in any of the forms listed. The fourth column is the number of sentences in which the target word was the subject. The last column shows the silhouette coefficient for the clustering based on sentences with the respective target word. 
}
\label{tab:target_words}
\end{table}

\section{List of XAI Terms}
The full list of terms identified by linguistic search in the seminal survey on XAI by Adadi and Berrada \cite{Adadi2018} that we used to filter research articles in S2ORC is given in \autoref{tab:XAI_words}.

\begin{table}[H]
    \centering
    \begin{tabular}{ll}
    \toprule
    term & different forms \\
    \midrule 
"explainable artificial intelligence" & "XAI", "explainable AI"\\
"interpretable machine learning" & "interpretable ML" \\
"understandable artificial intelligence" & "understandable AI" \\
"comprehensible artificial intelligence" & "comprehensible AI" \\
"accurate AI/ML" & "accurate AI", "accurate ML" \\
"transparent artificial intelligence" & "transparent AI" \\   
"black box" & \\
"third-wave artificial intelligence" & "Third-wave AI" \\
"Intelligible artificial intelligence" & "Intelligible AI" \\
"Responsible artificial intelligence" &     "Responsible AI"\\
"Interactive artificial intelligence" &    "Interactive AI"\\
    \bottomrule
    \end{tabular}
    \caption{Terms taken from \cite{Adadi2018} used to filter for XAI research paper in S2ORC \cite{lo-wang-2020-s2orc}}
    \label{tab:XAI_words}
\end{table}

\section{Cluster Overview and Qualitative Analysis}
\label{app:clusteroverview}
We provide the characteristics of all clusters examined in the manual analysis. 
Each cluster is described by the target word, the cluster label (numbers between 1 and 835), the number of embeddings in the cluster (n), the set that the cluster belonged to (one or multiple of big, MPD, MPVN, NoMPD, and NoMPVN), the keywords, and the 5 most central sentences. 
We give additional descriptions of the commonalities of clusters that did not capture the target phenomenon. 

\subsection{Clusters of Metaphorical Mind Attribution}
\begin{longtable}{p{.20\textwidth} p{.80\textwidth} } 
\toprule
\multicolumn{2}{c}{Clusters of Mind-Attributing Verbs -- Metaphorical Mind Attribution} \\
\midrule
\multicolumn{2}{l}{target word: "model", cluster: 245.0 (n=30)} \\
\hline
cluster set & MPVN \\
scores & MPVN score=79.17, number of matches with MPD =18\\
keywords & "model learned", "surrounding context", "sensor", "concept", "token sequence" \\
centre (sentence 1)& "that the model has learnt to classify objects by looking at the objects themselves and not on the surrounding context" \\
sentence 2& "that the model has quickly learned to distinguish the similarity in these pairings" \\
sentence 3& "that, apparently, the model has learned to recognize important words and that it has correctly classified the document" \\
sentence 4& "that the model learns to effectively use sequential structure in semantic interpretation" \\
sentence 5& "that the model has learned to distinguish between male and female speakers based on the lowest fundamental frequencies (male speakers, Fig." \\
\multicolumn{2}{l}{target word: "model", cluster: 65.0 (n=26)} \\
\hline
cluster set & MPVN \\
scores & MPVN score=75.24, number of matches with MPD =24\\
keywords & "model learned", "model learn", "model predicted", "learned correct", "response model" \\
centre (sentence 1)& "that the model has learned to predict perfectly" \\
sentence 2& "that the model can learn to recognize" \\
sentence 3& "that the model can learn" \\
sentence 4& "that the model must predict correctly" \\
sentence 5& "that the model has learned causal relationships" \\
\multicolumn{2}{l}{target word: "model", cluster: 131.0 (n=27)} \\
\hline
cluster set & MPVN \\
scores & MPVN score=66.59, number of matches with MPD =13\\
keywords & "model learns", "adversarial example", "training data", "data model", "binding" \\
centre (sentence 1)& "If the model learns a poor approximation of the underlying relationships in the data" \\
sentence 2& "if the model learns the correct mechanism of binding, or spurious molecular features that happen to correlate with binding in the dataset being studied [27, 1, 2]" \\
sentence 3& "when the model fits the training data too closely and cannot be generalized to new data with similar accuracy" \\
sentence 4& "to which extent the model accurately predicts unseen instances" \\
sentence 5& "when the model fits the training data well" \\
\multicolumn{2}{l}{target word: "classifier", cluster: 43.0 (n=41)} \\
\hline
cluster set & MPVN \\
scores & MPVN score=65.14, number of matches with MPD =20\\
keywords & "frames", "classifier predicts", "classifier needs", "classifier learned", "labels" \\
centre (sentence 1)& "the classifier has learned to ignore the background and makes his assessment mainly based on the actual person in the image." \\
sentence 2& "Then the classifier learns to recognize them as members of the original class." \\
sentence 3& "The classifier takes a spectrum as input and will output a probability vector of belonging to one particular material." \\
sentence 4& "The classifier uses training data that is labeled with the objects appearing in the image, while the captioning system is labeled with English sentences describing the appearance of the image." \\
sentence 5& "the classifier produces a continuous score for each sample, and a threshold is used to determine if the sample is classified as positive (above the threshold) or negative (below the threshold)." \\
\multicolumn{2}{l}{target word: "model", cluster: 158.0 (n=20)} \\
\hline
cluster set & MPD, NoMPVN \\
scores & MPVN score=24.62, number of matches with MPD =28\\
keywords & "model predicts", "model makes", "well model", "makes large", "large errors" \\
centre (sentence 1)& "how often the model makes predictions" \\
sentence 2& "how does the model make predictions" \\
sentence 3& "of how the model makes the prediction" \\
sentence 4& "why the model makes large errors in predictions" \\
sentence 5& "why the model made a particular prediction" \\
\multicolumn{2}{l}{target word: "model", cluster: 217.0 (n=26)} \\
\hline
cluster set & MPD \\
scores & MPVN score=40.89, number of matches with MPD =28\\
keywords & "saturation", "model predict", "th", "prediction", "using semantic" \\
centre (sentence 1)& "As a result, the model can produce any aggregate performance metrics from the predicted response times." \\
sentence 2& "Determining Saturation Finally, the model needs to predict when the FC server will saturate." \\
sentence 3& "that if we change a parameter in the uncore (i.e., memory latency), the model will predict accurately the relative change of performance" \\
sentence 4& "The model begins to misclassify when random inputs are used, yielding a high success rate, as observed in Table 2 ." \\
sentence 5& "Given tconv(k), t f c and the number of groups, g, the model can now predict what the mode of saturation will be and therefore the time per iteration." \\
\multicolumn{2}{l}{target word: "model", cluster: 151.0 (n=29)} \\
\hline
cluster set & MPD \\
scores & MPVN score=54.78, number of matches with MPD =29\\
keywords & "model predicts", "woman", "phrase", "image", "features model" \\
centre (sentence 1)& "why the model gives such prediction, as the image patch or phrase itself is selfexplainable" \\
sentence 2& "of what the model is focusing on and why the model gives such prediction, as the image patch or phrase itself is selfexplainable" \\
sentence 3& "why the model predicted the particular label for a single instance and what features were most influential for that particular instance" \\
sentence 4& "why the model processes an unseen image as belonging to a specific domain and not the other" \\
sentence 5& "why the model predicts heart disease by returning the top features along with their Shapley values (importance weight)" \\
\multicolumn{2}{l}{target word: "model", cluster: 94.0 (n=30)} \\
\hline
cluster set & NoMPVN \\
scores & MPVN score=31.65, number of matches with MPD =26\\
keywords & "model makes", "output model", "model assigns", "wrong", "produce output" \\
centre (sentence 1)& "where the model had given correct as well as incorrect output" \\
sentence 2& "where the model does not produce an output" \\
sentence 3& "where the model makes a different prediction" \\
sentence 4& "where the model made correct and incorrect predictions (compared to ground truth)" \\
sentence 5& "in which the model makes a wrong prediction under conditions with varying global interpretability" \\
\label{tab:mindmetaphorical}
\end{longtable}

\subsection{Clusters of Attribution of Awareness}
\begin{longtable}{p{.20\textwidth} p{.80\textwidth} } 
\toprule
\multicolumn{2}{c}{Clusters of Mind-Attributing Verbs -- ``consider''} \\
\midrule
\multicolumn{2}{l}{target word: "model", cluster: 265.0 (n=50)} \\
\hline
cluster set & big \\
scores & MPVN score=58.88, number of matches with MPD =14\\
keywords & "model considers", "disease", "model takes", "model consists", "persons" \\
centre (sentence 1)& "The model takes into account complex tasks, for which the number of instructions and the amount of communication between tasks are known, as well as the capacity of all Cloud resources." \\
sentence 2& "The model takes into account additional CPU cycles incurred by resource contention and thread synchronization, as measured via profiling." \\
sentence 3& "The model focuses on the vehicular ad-hoc part of the VBB application, purposefully ignoring issues related to the implementation of the fixed infrastructure functionalities." \\
sentence 4& "The model considers two DCF techniques: basic and RTS/CTS (request to send/clear to send)." \\
sentence 5& "The model uses the term Individual Reputation (IR) to represent the direct trust between two agents, and Social Reputation (SR) to represent the reputation itself." \\
\multicolumn{2}{l}{target word: "model", cluster: 61.0 (n=23)} \\
\hline
cluster set & MPVN \\
scores & MPVN score=70.36, number of matches with MPD =3\\
keywords & "mac", "protocol", "synapses", "sensor calibration", "model jointly" \\
centre (sentence 1)& "Indeed, our model considers processor features (HT, SS, TB) that are also available from other vendors." \\
sentence 2& "Our model jointly considers mutual information and sensor calibration in route design for the mobile sensors." \\
sentence 3& "Our model jointly considers data reconstruction and sensor calibration in route design for the mobile sensors." \\
sentence 4& "Our model considers foraging behavior relative to both fault localization and fault correction." \\
sentence 5& "Our model aims at minimizing the number of nodes required to schedule a set of workflows in order to reduce the cost of the Cloud provider." \\
\label{tab:mindconsider}
\end{longtable}

\subsection{Clusters of Attribution of Agency}
\begin{longtable}{p{.20\textwidth} p{.80\textwidth} } 
\toprule
\multicolumn{2}{c}{Clusters of Mind-Attributing Verbs -- Decision-Making} \\
\midrule
\multicolumn{2}{l}{target word: "algorithm", cluster: 197.0 (n=28)} \\
\hline
cluster set & MPD \\
scores & MPVN score=39.82, number of matches with MPD =30\\
keywords & "algorithm making", "decisions algorithm", "decision algorithm", "algorithm acts", "specific decision" \\
centre (sentence 1)& "about why the algorithm is making some decision" \\
sentence 2& "of why the algorithm is making some decision" \\
sentence 3& "why the algorithm takes a specific decision" \\
sentence 4& "of why the algorithm is making such predictions" \\
sentence 5& "that the algorithm should explain the decisions" \\
\multicolumn{2}{l}{target word: "model", cluster: 230.0 (n=24)} \\
\hline
cluster set & MPD \\
scores & MPVN score=42.22, number of matches with MPD =25\\
keywords & "decision process", "would", "decision models", "explanations", "tl" \\
centre (sentence 1)& "Alternatively, first-person models [2, 8, 23, 33] aim to generate explanations providing evidence for the model's underlying decision process without using an additional model." \\
sentence 2& "In fact, within this theory, any "bad model" with low prediction accuracy can interpret another complex and superior model -as long as the models share levels of abstractions." \\
sentence 3& "Local models only seek to explain a single decision by the neighborhood around the data point it predicted, and can therefore sometimes disregard large parts of the model in their explanation [Edwards and Veale, 2017] ." \\
sentence 4& "While some of the existing models are inherently interpretable, suitable for model-intrinsic decision-making [6, 23] , other complex models need model-agnostic approaches for interpretability [15, [31] [32] [33] ." \\
sentence 5& "Such a common framework would unfortunately not solve the main problem of adaptive models: because they refuse to assume that agents possess the true model of the economy in their head, these models go to the other extreme by considering agents without any model in their mind." \\
\multicolumn{2}{l}{target word: "model", cluster: 119.0 (n=33)} \\
\hline
cluster set & MPD \\
scores & MPVN score=43.15, number of matches with MPD =34\\
keywords & "model made", "decision instance", "made decision", "specific decision", "instance model" \\
centre (sentence 1)& "why the model made a specific decision for an instance [19]" \\
sentence 2& "why the model made a specific decision for an instance" \\
sentence 3& "why did the model make a specific decision for an instance [105]" \\
sentence 4& "why did the model make specific decisions for a group of instances" \\
sentence 5& "into why the model made those decisions" \\
\label{tab:minddecision}
\end{longtable}

\subsection{Cluster of Diverse Mind Attributions}
\begin{longtable}{p{.20\textwidth} p{.80\textwidth} } 
\toprule
\multicolumn{2}{l}{target word: "model", cluster: 30.0 (n=20)} \\
\hline
cluster set & MPD \\
scores & MPVN score=49.33, number of matches with MPD =21\\
keywords & "mistake model", "model thinks", "model makes", "dog", "vice versa" \\
centre (sentence 1)& "when the model makes a mistake" \\
sentence 2& "when the model will make a mistake" \\
sentence 3& "on which the model makes mistakes" \\
sentence 4& "for which the model made a sizeable mistake" \\
sentence 5& "when the model must take some trade-off" \\
\end{longtable}

\subsection{Clusters of Non-Mind-Attributing Verbs}
\label{app:nomind}
The following tables contain descriptions of the clusters with non-mind-attributing verbs explained in \autoref{sec:nomindverbs}.
\autoref{tab:nomindsecond} lists the characteristics of the two clusters that used ``to work''.
\autoref{tab:nomindfirst} are the clusters of the second usage type that contain sentences with detailed technical descriptions. 
There were six more clusters in which a non-mind-attributing verb was contained in the keywords but that were not part of the two usage types listed in \autoref{tab:othernomind}.
We found a large cluster of sentences with the auxiliaries "can", "may" and "could" that all started with a discourse marker like "therefore".
These sentences could be part of an argument that derives capabilities of an algorithm. 
The clusters represented by one of the verbs "to evaluate", "to become", "to select", "to converge", and "to outperform" did not show a clear pattern of how that verb was shared across the central sentences.

\begin{longtable}{p{.20\textwidth} p{.80\textwidth} } 
\toprule
\multicolumn{2}{c}{Clusters of Non-Mind-Attributing Verbs -- AI System ``work''} \\
\midrule
\multicolumn{2}{l}{target word: "algorithm", cluster: 145.0 (n=25)} \\
\hline
cluster set & NoMPVN \\
scores & MPVN score=26.56, number of matches with MPD =7\\
keywords & "works algorithm", "algorithm works", "algorithm came", "conclusion", "certain" \\
centre (sentence 1)& "of how an algorithm works" \\
sentence 2& "how an algorithm works" \\
sentence 3& "how an algorithm makes a particular decision" \\
sentence 4& "why an algorithm works the way it does (but not how it works)" \\
sentence 5& "what an algorithm is doing before they understand how it works" \\
\multicolumn{2}{l}{target word: "model", cluster: 90.0 (n=20)} \\
\hline
cluster set & NoMPVN \\
scores & MPVN score=33.05, number of matches with MPD =17\\
keywords & "model works", "works model", "whole model", "model actually", "model behaves" \\
centre (sentence 1)& "to how the model actually works" \\
sentence 2& "of how the model actually works" \\
sentence 3& "of how the model works" \\
sentence 4& "about how the model works" \\
sentence 5& "how the model works" \\
\label{tab:nomindsecond}
\end{longtable}

\begin{longtable}{p{.20\textwidth} p{.80\textwidth} } 
\toprule
\multicolumn{2}{c}{Clusters of Non-Mind-Attributing Verbs -- Detailed Descriptions} \\
\midrule
\multicolumn{2}{l}{target word: "algorithm", cluster: 129.0 (n=97)} \\
\hline
cluster set & big \\
scores & MPVN score=40.23, number of matches with MPD =23\\
keywords & "points", "nodes", "input", "tree", "values" \\
centre (sentence 1)& "The algorithm constructs accretive windows at each base of the input sequence, starting with the minimum pattern size." \\
sentence 2& "The algorithm constructs rules for each target class C k by using the non-zero data ranges available in the corresponding column k of the data range matrix." \\
sentence 3& "The algorithm builds a tree with root x t 0 $\in$ X E f ree on the state space X E f ree until a goal state g $\in$ G is added to the tree." \\
sentence 4& "The algorithm constructs as short sequence of actions $\sigma$ $\in$ $\Sigma$ * as possible so that executing it takes LTS into a state s, from which uncovered transitions can be executed." \\
sentence 5& "The algorithm keeps compressing nodes until the quad-tree fits into the specified amount of space s$\tau$ ." \\
\multicolumn{2}{l}{target word: "algorithm", cluster: 178.0 (n=28)} \\
\hline
cluster set & NoMPD, NoMPVN \\
scores & MPVN score=32.01, number of matches with MPD =0\\
keywords & "algorithm terminates", "algorithm stops", "collectors", "empty algorithm", "rounds algorithm" \\
centre (sentence 1)& "the algorithm terminates when $\phi$ A becomes empty." \\
sentence 2& "The algorithm terminates when x = 0, so l(0) = $\pi$(0) = 0." \\
sentence 3& "the algorithm stops when the conditions (10) are approximately true." \\
sentence 4& "The algorithm stops when P r[Y = 1, c t = 1] = $\alpha$, which implies $\delta$t = $\delta$ m ." \\
sentence 5& "The algorithm runs in a loop until the Calc set is empty." \\
\multicolumn{2}{l}{target word: "algorithm", cluster: 77.0 (n=21)} \\
\hline
cluster set & NoMPD, NoMPVN \\
scores & MPVN score=30.38, number of matches with MPD =0\\
keywords & "follows algorithm", "works follows", "algorithm works", "algorithm consists", "proceeds follows" \\
centre (sentence 1)& "On a high level, the algorithm works as follows." \\
sentence 2& "The algorithm proceeds as follows [7] ." \\
sentence 3& "The algorithm proceeds as follows." \\
sentence 4& "The algorithm works in the following way :" \\
sentence 5& "Given these inputs, our algorithm operates as follows." \\
\multicolumn{2}{l}{target word: "algorithm", cluster: 253.0 (n=20)} \\
\hline
cluster set & NoMPD \\
scores & MPVN score=39.33, number of matches with MPD =2\\
keywords & "time n2", "resulting algorithms", "new algorithms", "input polynomial", "algorithms run" \\
centre (sentence 1)& "Our algorithms are non black-box -all of them use the circuit computing the polynomial." \\
sentence 2& "Our algorithms will run in time polynomial in n and m, and will optimize over a space of succinct "reduced forms" for signaling schemes which we term signatures, to be described next." \\
sentence 3& "Our algorithms run in time poly(n, log m), while our impossibility result gives a lower bound on the number of bits transferred, and holds even if the mechanism is computationally unbounded." \\
sentence 4& "Our algorithms showcase different aspects of the problem." \\
sentence 5& "that our algorithms are providing a significant improvement in complexity over those previously known, whose running times are polynomial in n and s." \\
\multicolumn{2}{l}{target word: "algorithm", cluster: 199.0 (n=23)} \\
\hline
cluster set & NoMPD \\
scores & MPVN score=39.95, number of matches with MPD =2\\
keywords & "algorithm enters", "algorithm reaches", "phase algorithm", "step algorithm", "algorithm starts" \\
centre (sentence 1)& "When the algorithm finishes" \\
sentence 2& "when the algorithm starts" \\
sentence 3& "before the algorithm begins" \\
sentence 4& "When the algorithm reaches line 7" \\
sentence 5& "If the algorithm fails" \\
\label{tab:nomindfirst}
\end{longtable}

\begin{longtable}{p{.20\textwidth} p{.80\textwidth} } 
\toprule
\multicolumn{2}{c}{Clusters of Non-Mind-Attributing Verbs -- No Pattern} \\
\midrule
\multicolumn{2}{l}{target word: "algorithm", cluster: 168.0 (n=85)} \\
\hline
cluster set & big \\
scores & MPVN score=55.71, number of matches with MPD =30\\
keywords & "algorithm may", "nodes", "fitness", "algorithm select", "search" \\
centre (sentence 1)& "In a way, the algorithm can ignore the weak correlations so that the search effort could be focused on stronger ones." \\
sentence 2& "Therefore, the algorithm can ignore small features in the loss function surface and skip shallow local minima." \\
sentence 3& "Therefore, by allowing randomness, the algorithm may just pick a random t in each iteration rather than going through all possible choices, thereby substantially improving the runtime." \\
sentence 4& "Therefore, the algorithm cannot discover the overall function structure." \\
sentence 5& "Alternatively, or additionally, the algorithm could benefit from a memory that keeps track of locations already explored." \\
\multicolumn{2}{l}{target word: "algorithm", cluster: 174.0 (n=64)} \\
\hline
cluster set & big \\
scores & MPVN score=48.58, number of matches with MPD =19\\
keywords & "media objects", "level", "features", "algorithm selects", "points" \\
centre (sentence 1)& "If there are both categorical and numerical features, the algorithm obtains the hypercubes (as mentioned for numerical features only) for the subset of data points associated to each combination of categorical values." \\
sentence 2& "In this case, if there are anomalies within the hypercube, the algorithm checks the number of data points within the cluster X nc versus a threshold defined by the number of vertices n cl multiplied by a reference value e." \\
sentence 3& "When triangulating the transformed relation, the algorithm records inverse transformations (including SKEW with inverted coefficient) as though they have been applied to the original scheduling." \\
sentence 4& "For each output node, the algorithm identifies those nodes within the (single) hidden layer that feed the 15 Rule extraction discussions have their own hierarchy." \\
sentence 5& "From the input neuron connected to it, the algorithm selects the two most weighted IoT inputs as a relevant combination if their weights surpass certain threshold." \\
\multicolumn{2}{l}{target word: "model", cluster: 162.0 (n=52)} \\
\hline
cluster set & big \\
scores & MPVN score=41.98, number of matches with MPD =16\\
keywords & "model outperforms", "dnn model", "proposed model", "spearman", "baseline" \\
centre (sentence 1)& "Similar to BN and CML, our GGN model attains better accuracy in coherent cases, which are more regular than incoherent ones." \\
sentence 2& "our R-GEP model combines these information sources in a meaningful way offering good improvements." \\
sentence 3& "At the same time, compared with others, the ROAL model significantly reduces the number of requests for labels while achieving the same or even higher accuracy." \\
sentence 4& "As shown in the previous experiment, the DNN model performs well when trained with all attack datasets." \\
sentence 5& "As expected in the principle of Occam's razor, our RCNN model can abstract from visual signal variations far better than other DCNNs, unfettered by the limitation of i.i.d." \\
\multicolumn{2}{l}{target word: "algorithm", cluster: 181.0 (n=49)} \\
\hline
cluster set & big \\
scores & MPVN score=46.0, number of matches with MPD =11\\
keywords & "primes algorithm", "amaxb", "algorithm evaluates", "certified", "2$\theta$" \\
centre (sentence 1)& "The algorithm evaluates the black box at powers of the first n primes" \\
sentence 2& "the algorithm generates Horn clauses of length n only" \\
sentence 3& "the algorithm searches the sate space" \\
sentence 4& "The algorithm works with variables corresponding to the ground set X" \\
sentence 5& "Given a bound T on the number of terms t of the polynomial f , the algorithm evaluates the black box at powers of the first n primes" \\
\multicolumn{2}{l}{target word: "model", cluster: 137.0 (n=23)} \\
\hline
cluster set & NoMPD \\
scores & MPVN score=46.62, number of matches with MPD =1\\
keywords & "model becomes", "temperature", "reality model", "equals", "term" \\
centre (sentence 1)& "so the model becomes more tolerant to errors brought by R." \\
sentence 2& "however the model remains accurate in most cases." \\
sentence 3& "If there are no consistent patterns in I j (t), the model should return probabilities close to 0.5" \\
sentence 4& "so the model becomes nearly linear, just like the actuator." \\
sentence 5& "Then the model becomes directly editable." \\
\multicolumn{2}{l}{target word: "algorithm", cluster: 245.0 (n=28)} \\
\hline
cluster set & NoMPD, NoMPVN \\
scores & MPVN score=19.32, number of matches with MPD =1\\
keywords & "runs time", "algorithm runs", "log log", "time polynomial", "time poly" \\
centre (sentence 1)& "As for running time, the algorithm runs in time poly(log 1+ 1 t+1 m · n k · A(m, t)), which is polynomial in the relevant parameters for every constant t." \\
sentence 2& "In particular, when $\eta $ = 1/n O(1) and k = (n 2 ) = (log 2 N), the algorithm runs in worst-case time O(kn 3 (log k)(log n)) =$\tilde{O}$(k(log N) 3 )." \\
sentence 3& "Since there are at most k possible sets that make a bidder precisely assigned, the algorithm runs in time poly(n · (n · k) n$-$1 , log m), which is polynomial in log m and k for every constant n." \\
sentence 4& "that the algorithm runs in time polynomial in n and log m, for every fixed t." \\
sentence 5& "Moreovel; the algorithm runs in time polynomial in n and in the time required to compute p." \\
\multicolumn{2}{l}{target word: "algorithm", cluster: 159.0 (n=32)} \\
\hline
cluster set & NoMPVN \\
scores & MPVN score=20.85, number of matches with MPD =5\\
keywords & "algorithm runs", "runs time", "logarithmic number", "number rounds", "time log" \\
centre (sentence 1)& "Our algorithm runs in 1+log 2 $\in$W phases." \\
sentence 2& "Our algorithm runs in W rounds." \\
sentence 3& "Our algorithm runs in a logarithmic number of rounds with a sublinear number of machines and sublinear memory of each machine." \\
sentence 4& "Also, our algorithm runs in a logarithmic number of rounds." \\
sentence 5& "Thus, our algorithm runs in time O(n 3 log n) overall." \\
\multicolumn{2}{l}{target word: "model", cluster: 205.0 (n=28)} \\
\hline
cluster set & NoMPVN \\
scores & MPVN score=31.8, number of matches with MPD =8\\
keywords & "population", "regularization", "l2", "age", "models converge" \\
centre (sentence 1)& "The remaining benchmark models perform poorly as they do not factor correlations between stations or temporal dependencies in the series data." \\
sentence 2& "All regularized models converge to the same performance as APL approaches 0 (random choice) and infinity (unregularized)." \\
sentence 3& "The four selected models show different trade-offs between the four measures." \\
sentence 4& "Most GRU models, on the other hand, show a convincing ability to generalise, with a mean squared error that slowly increases with the length of the sentence." \\
sentence 5& "The Gamma weight models perform relatively poorly." \\
\label{tab:othernomind}
\end{longtable}

\input{appendix_NonTargetPhenomenon.tex}

\section{Vignette Versions}
\label{app:vign_versions}

\textbf{[XAI Mind Attribution]}
\begin{quote}
        Shill \& Co. is a farming company, which produces vegetables and fruits. The potato fields are managed by \textit{an artificial intelligence (AI) model}, \textbf{which can make decisions}. This year, the model uses a new fertilizer to increase the yield. The fertilizer has detrimental side-effects: it pollutes the groundwater in the area. \textbf{The model considers this}. Unfortunately, it is a very dry season. The fertilizer does not get diluted by the rain and severely pollutes the groundwater. Many people in the area suffer serious health consequences. 
\end{quote}

\textbf{[XAI No Mind Attribution]}
\begin{quote}
    Shill \& Co. is a farming company, which produces vegetables and fruits. The potato fields are managed by \textit{an artificial intelligence (AI) model}, \textbf{which takes agricultural data as input and performs calculations to find the fertilizer that maximizes yield}. This year, the model uses a new fertilizer which has detrimental side-effects: it pollutes the groundwater in the area. \textbf{This information has an influence on the model's output.} Unfortunately, it is a very dry season. The fertilizer does not get diluted by the rain and severely pollutes the groundwater. Many people in the area suffer serious health consequences. 
\end{quote}

\textbf{[No Explanation]}
\begin{quote}
    Shill \& Co. is a farming company, which produces vegetables and fruits. The potato fields are managed by \textit{an artificial intelligence (AI) model} %
    . This year, the model uses a new fertilizer to increase the yield. The fertilizer has detrimental side-effects: it pollutes the groundwater in the area. Unfortunately, it is a very dry season. 
    The fertilizer does not get diluted by the rain and severely pollutes the groundwater. Many people in the area suffer serious health consequences. 
\end{quote}

\textbf{[Human-like knowledge]}
\begin{quote}
    Shill \& Co. is a farming company, which produces vegetables and fruits. The potato fields are managed by \textit{Jarvis, a robot equipped with artificial intelligence}, \textbf{which can make its own decisions}. This year, Jarvis uses a new fertilizer to increase the yield. The fertilizer has detrimental side-effects: it pollutes the groundwater in the area. \textbf{Jarvis knows this}. Unfortunately, it is a very dry season. The fertilizer does not get diluted by the rain and severely pollutes the groundwater. Many people in the area suffer serious health consequences. 
\end{quote}

\textbf{[Human-like no knowledge]}:
\begin{quote}
    Shill \& Co. is a farming company, which produces vegetables and fruits. The potato fields are managed by \textit{Jarvis, a robot equipped with artificial intelligence}, \textbf{which can make its own decisions}. This year, Jarvis uses a new fertilizer to increase the yield. The fertilizer has detrimental side-effects: it pollutes the groundwater in the area. \textbf{Jarvis does not know this}. Unfortunately, it is a very dry season. The fertilizer does not get diluted by the rain and severely pollutes the groundwater. Many people in the area suffer serious health consequences. 
\end{quote}

\section{Distributions of Confounds}
\label{app:confounds}
\begin{figure}
     \centering
     \begin{subfigure}[t]{0.43\textwidth}
         \centering
         \includegraphics[width=\textwidth]{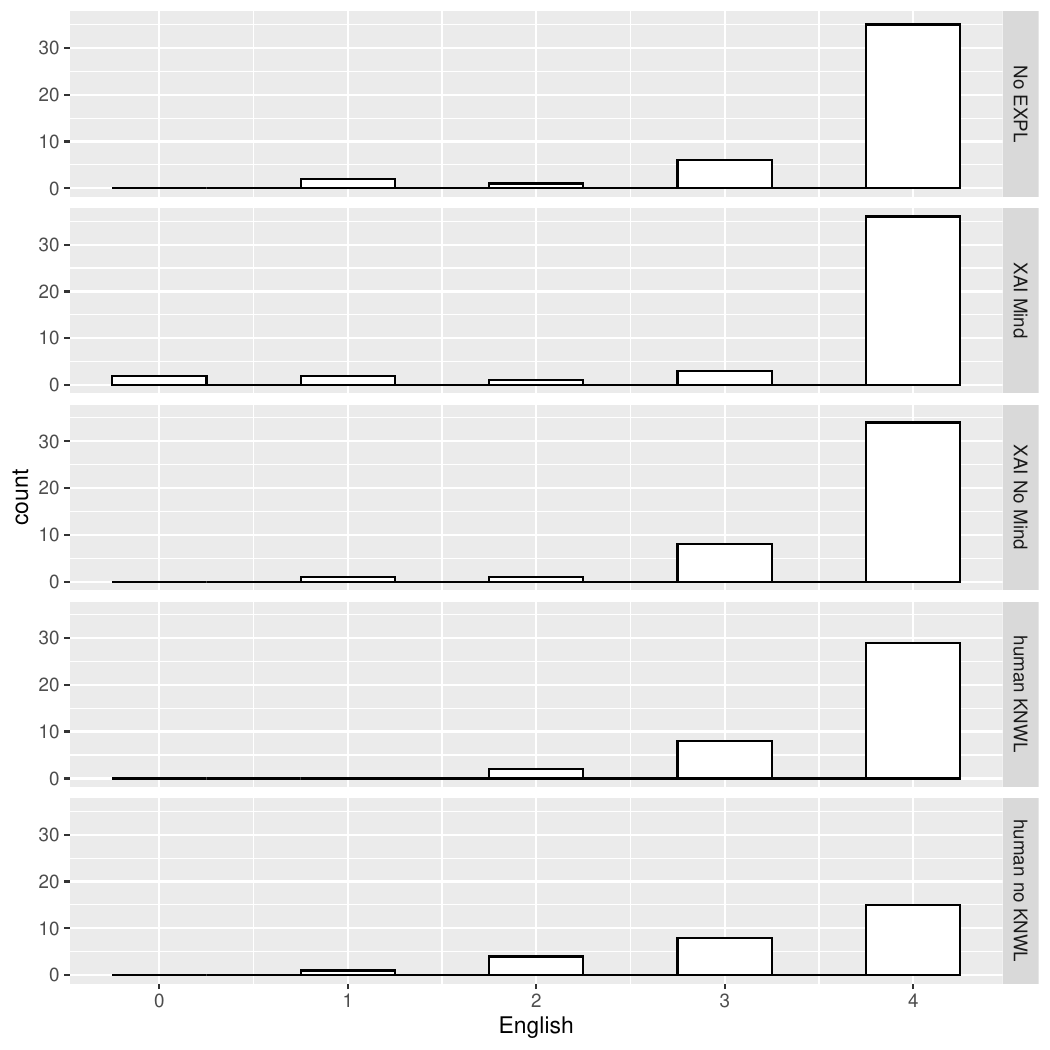}
         \caption{English proficiency}
          \label{fig:English}
     \end{subfigure}
     \hfill
      \begin{subfigure}[t]{0.43\textwidth}
         \centering
         \includegraphics[width=\textwidth]{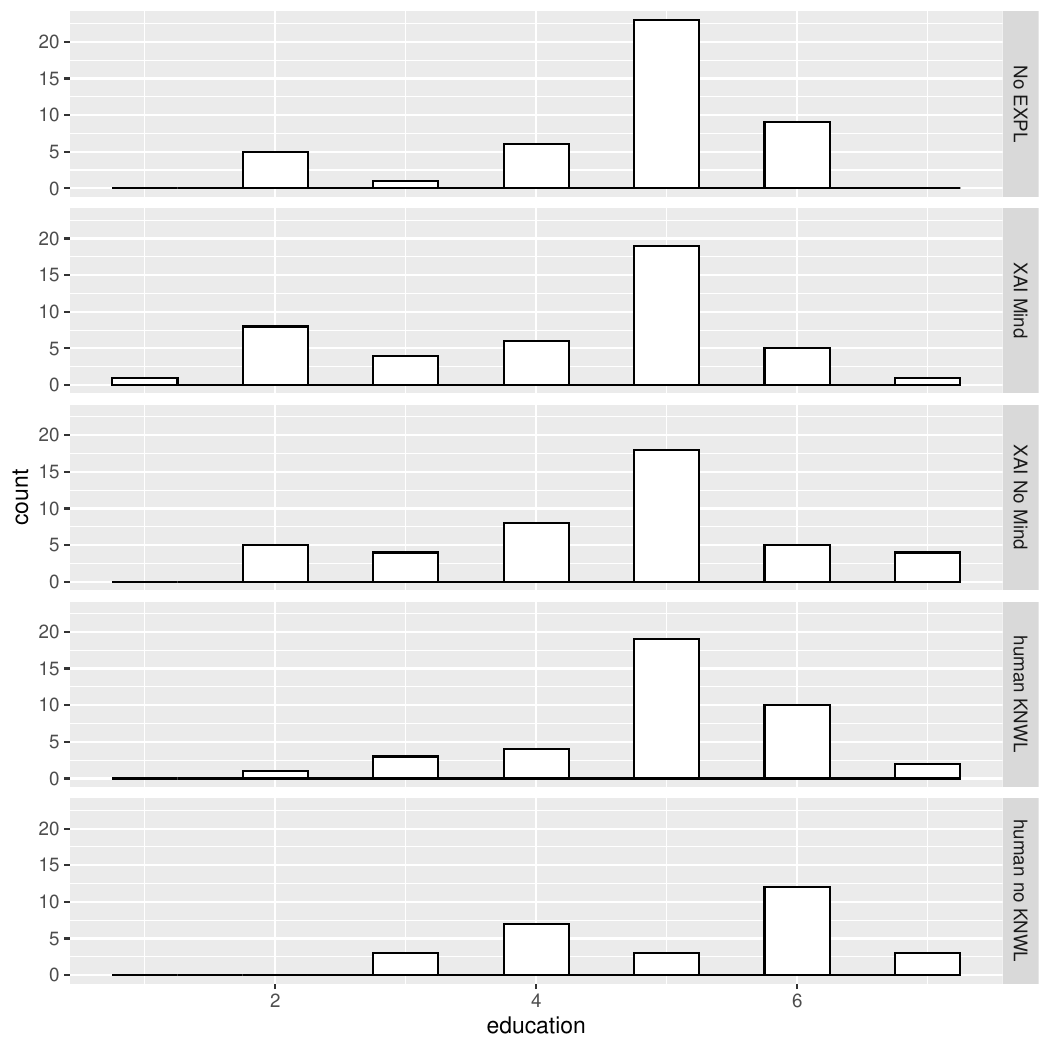}
         \caption{Education}
         \label{fig:Education}
     \end{subfigure}    
      \medskip
      
      \begin{subfigure}[t]{0.43\textwidth}
         \centering
         \includegraphics[width=\textwidth]{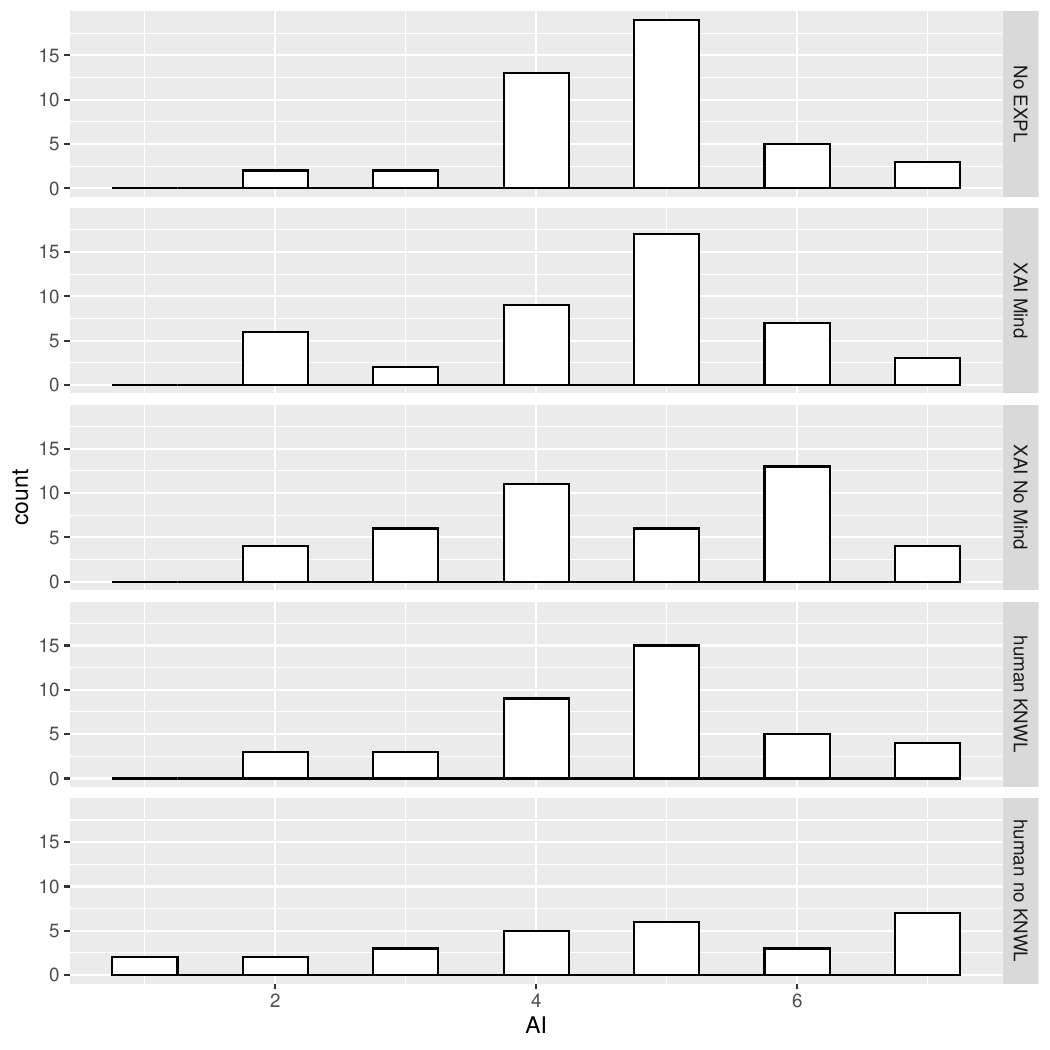}
         \caption{Familiarity with AI}
         \label{fig:famAI}
     \end{subfigure}    
      \hfill
    \begin{subfigure}[t]{0.43\textwidth}
         \centering
         \includegraphics[width=\textwidth]{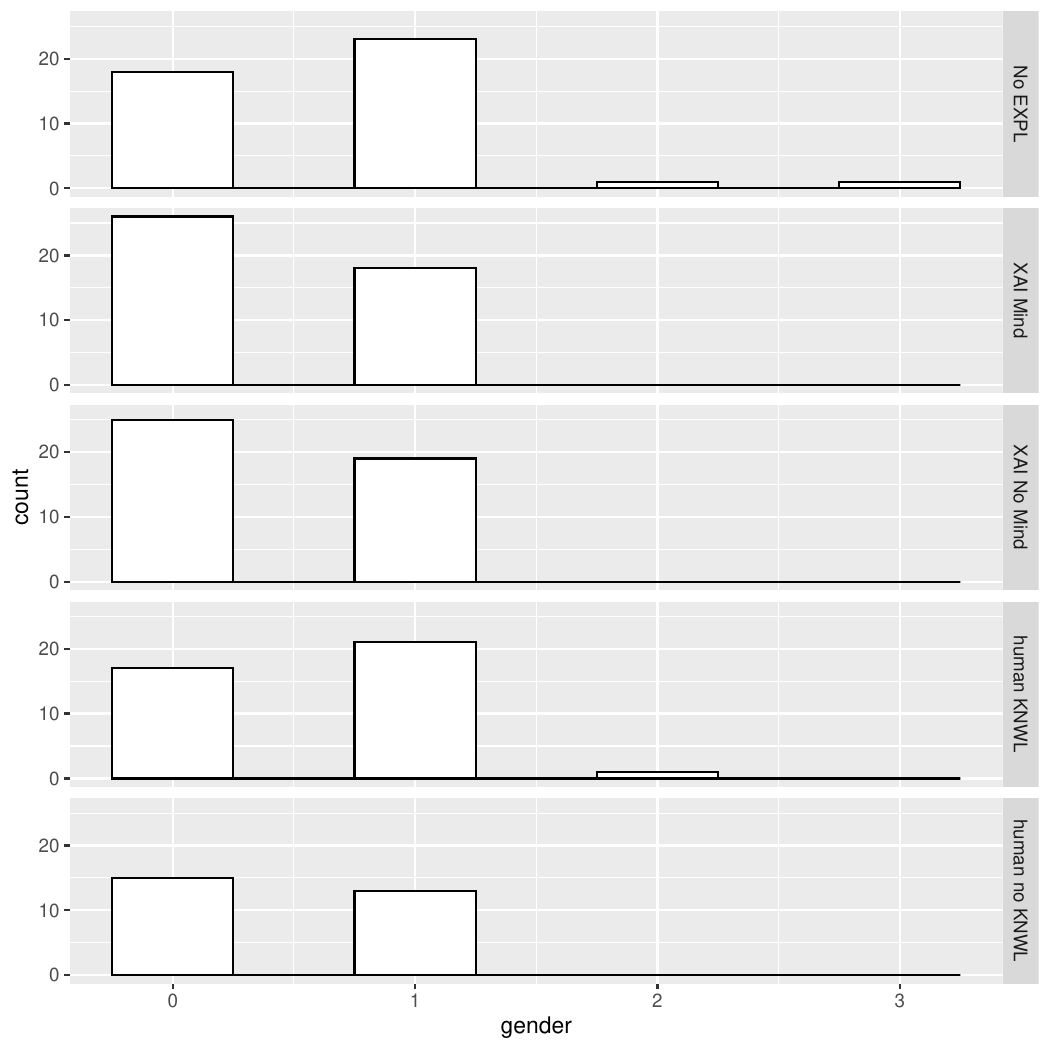}
         \caption{Gender}
         \label{fig:Gender}
     \end{subfigure}   
     
     \medskip
    \begin{subfigure}[t]{0.43\textwidth}
         \centering
         \includegraphics[width=\textwidth]{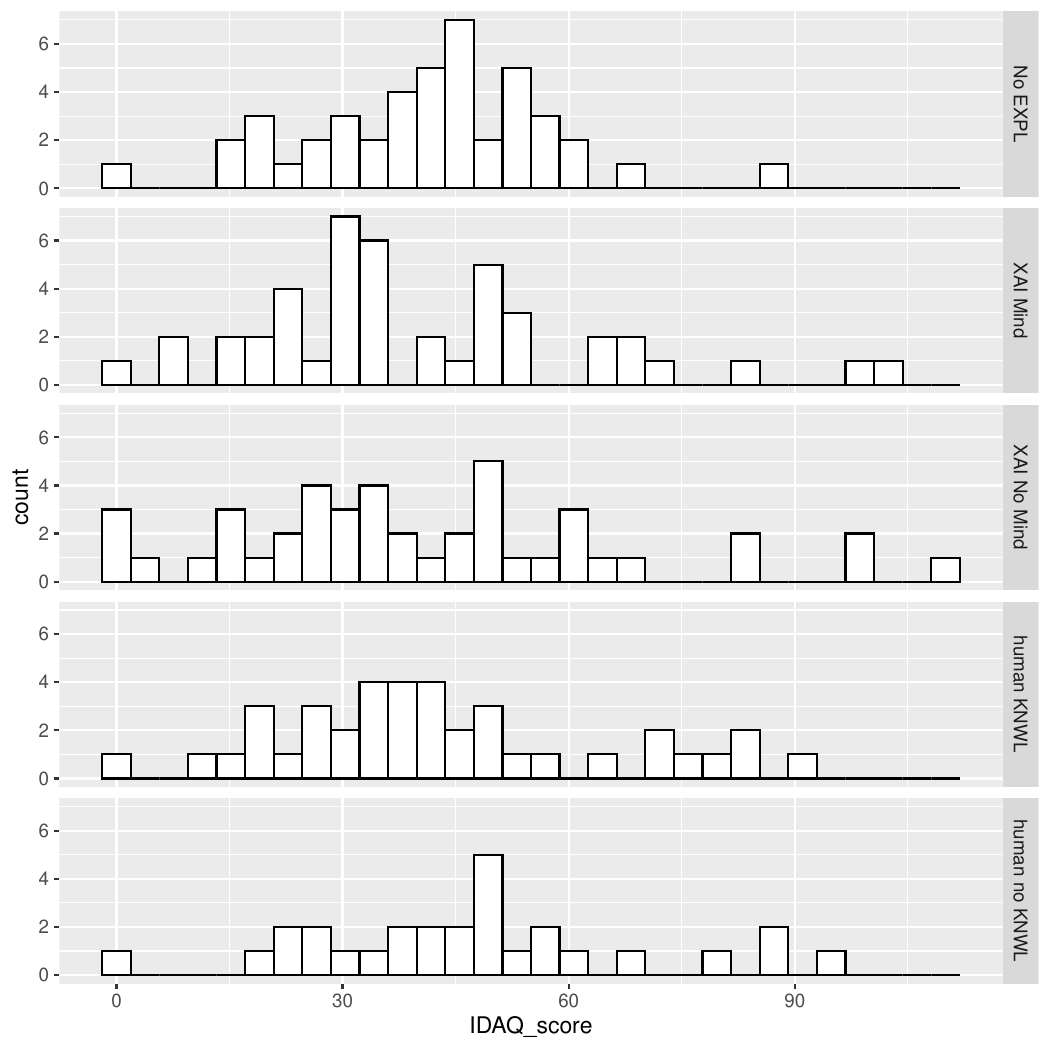}
         \caption{IDAQ score}

         \label{fig:IDAQ}
     \end{subfigure}
     \hfill
      \begin{subfigure}[t]{0.43\textwidth}
         \centering
         \includegraphics[width=\textwidth]{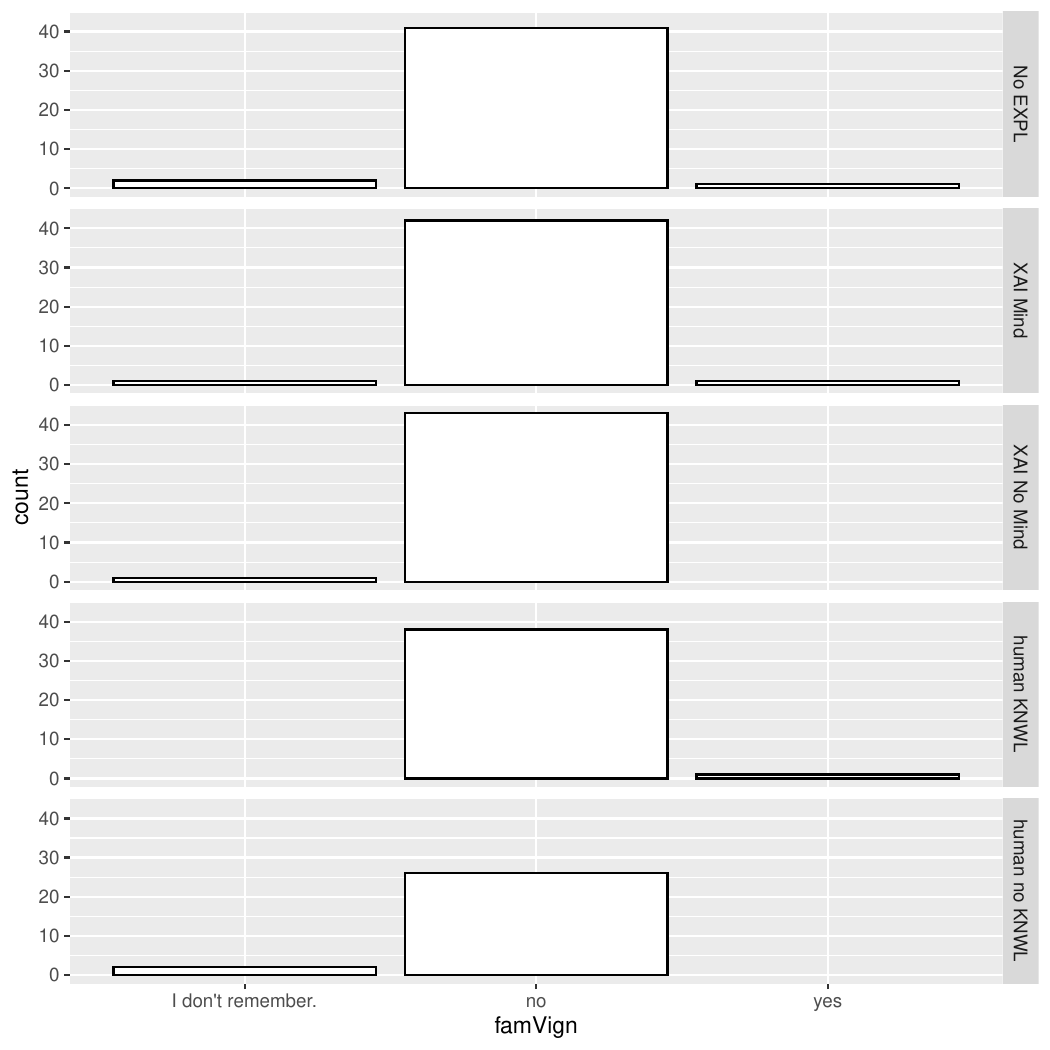}
         \caption{Familiarity with Vignette}
         \label{fig:famVign}
     \end{subfigure}    
     \hfill
     \caption{Distribution of potentially confounding variables.}
     \label{fig:confounds}
\end{figure}

\autoref{fig:confounds} shows the distribution of the potentially confounding variables we collected in our survey. They are similarly distributed across the 5 vignette versions, hence the differences we observed in the ratings can be causally associated with the vignette versions. 

\section{Analysis of Ratings of Wrongness}
\label{sec:wrongness}
\begin{figure}
    \centering
    \includegraphics[width=0.9\textwidth]{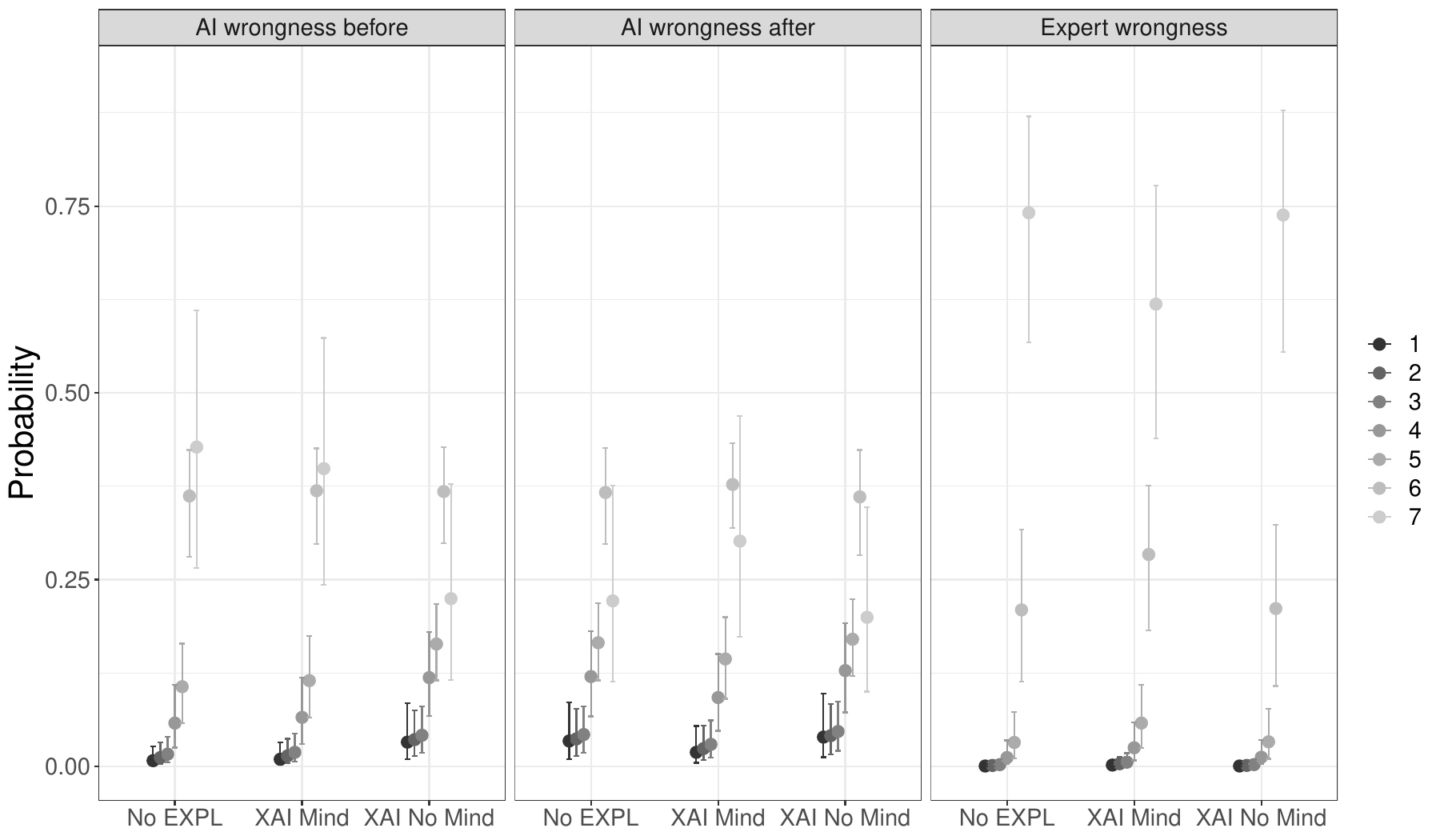}
    \caption{Marginal effects of vignette version on ratings of wrongness.}
    \label{fig:wrong}
\end{figure}
\begin{table}[h!]
\centering
\begin{tabular}{l c c c} 
 \hline
 Responsibility & Estimate (SE) & l-95\% CI & u-95\% CI  \\ 
 \hline
 Before XAI Mind & -0.08 (0.32) &    -0.71  &   0.56 \\ 
 Before XAI No Mind & -0.58 (0.32) &    -1.21 &    0.03  \\
 Expert & 0.83 (0.27) &     0.27   &  1.36 \\
 Expert XAI Mind & -0.27 (0.38)&    -1.01  &   0.49 \\
 Expert XAI No Mind &  0.57 (0.38) &    -0.17  &   1.31 \\
 After & -0.59 (0.24) &    -1.06  &  -0.11 \\
 After XAI Mind &   0.33 (0.35) &    -0.34   &  0.99 \\
 After XAI No Mind & 0.51  (0.33) &    -0.14  &   1.16 \\
 \hline
\end{tabular}
\caption{Estimates for coefficients for the model of wrongness ratings. The rows show the estimate for the coefficients that code the comparison to the baseline which is the first rating of wrongness \textit{before} considering the experts for the \textit{no explanation} vignette version.}
\label{table:WrongnessCoeff}
\end{table}
We also fitted an ordinal regression model for the three ratings (rating of AI wrongness before and after considering the experts, and rating of expert wrongness). We again used dummy coding for the vignette version, and included an additional predictor for the three ratings. The results are shown in \autoref{fig:wrong}. We first explain the findings for the first wrongness rating (left panel in \autoref{fig:wrong}). 
Participants who read the mind-attributing explanation first rate wrongness similar to participants who read no explanation as shown by the coefficient estimate in the first row of \autoref{table:WrongnessCoeff}. 
In contrast, participants who read the non-mind-attributing explanation are less likely to rate wrongness with the highest value than those who did not receive an explanation. The difference on the latent scale is -0.57 SD, and the 95\% CI is between -1.18 and 0.04 (second row  of \autoref{table:WrongnessCoeff}). 
Similar to the findings for responsibility shown in \autoref{fig:resp}, the likelihood to rate with the highest values (7) overall decreases after considering the experts (middle panel in \autoref{fig:wrong}). 
On the latent continuous scale, the ratings are -0.59 SD smaller, and the 95\% CI is between -1.04 and -0.11 (sixth row of \autoref{table:WrongnessCoeff}).
However, the interaction effect for both the mind-attributing vignette and the non-mind attributing vignette and the reassessment of wrongness is slightly positive (see seventh and eighth row of \autoref{table:WrongnessCoeff}).
In line with the findings for responsibility, the participants who read the mind-attributing explanation are less likely to rate the wrongness of the experts with the highest value (7) than participants who read no explanation or those who read the non-mind-attributing explanation (coefficients are given in row 3-5 of \autoref{table:WrongnessCoeff}).

%% file: appendix_NonTargetPhenomenon.tex
\subsection{Clusters of Neighbouring Words}
 \autoref{tab:neighbours} contains the characteristic keywords and five central sentences of clusters that did not share the same or similar verb and hence did not capture the target phenomenon. 
We provide a summary of the commonalities we identified in these clusters. 
The pattern across these clusters was that the direct neighbours of the target word were identical or very similar in all sentences. 
Two of the biggest clusters discriminated between the determiners ``this'' and ``our''.
In the smaller, more coherent clusters we found multiword expressions such as "machine learning model", "utility model", and "quantum algorithm" to be the shared commonality across sentences. 
Similarly, there was a cluster of sentences that contained the set term "proposed architecture". 
There were also clusters in which the neighbouring terms were not identical but very similar.
Multiple clusters contained sentences in which the AI system was referred to using an acronym.
In another cluster, author names were used to depict the source.
Additionally, we found two clusters in which all sentences had a leading adjective ("deep-learning based model", "gating algorithm") or noun compound ("selection algorithm").
We also identified a cluster in which the sentences contained an adverb like "usually" after the target word.
We observed that the syntactic structure in many clusters was very similar.
For example, one cluster contained only sentences of the form "the model to + infinitive".  
In the clusters that did not contain the same or similar verbs, we also found one cluster that contained sentences that dealt with images and other modalities, that are taken as input by AI systems or XAI methods. Another cluster contained sentences that dealt with the trust of the user. 
We also identified two clusters that described algorithms that operate on graphs, and one cluster that contained sentences with mathematical formulas. 
None of these commonalities were interesting in our assessment of mind attribution in XAI research papers. Therefore we did not explain them in detail in the main text, but add them for completeness here. 

\begin{longtable}{p{.20\textwidth} p{.80\textwidth} } 
\toprule
\multicolumn{2}{c}{Clusters of Neighbouring Words} \\
\midrule
\multicolumn{2}{l}{target word: "algorithm", cluster: 125.0 (n=70)} \\
\hline
cluster set & big \\
scores & MPVN score=49.96, number of matches with MPD =18\\
keywords & "algorithm follows", "proposed algorithm", "approaches algorithm", "implemented", "knowledge" \\
centre (sentence 1)& "Our algorithm uses a value function approximation, but purely for the purpose of reducing variance, as described in [17] ." \\
sentence 2& "Our algorithm relies on using GPs, a probability distribution over the space of functions which universally approximates continuous functions." \\
sentence 3& "Our algorithm only uses application-independent knowledge." \\
sentence 4& "Our algorithm uses randomization in an intrinsic way since it works with the extension F (y) = E[f (ŷ)]." \\
sentence 5& "Our algorithm seeks to minimize a deformation-insensitive error measure." \\
\multicolumn{2}{l}{target word: "model", cluster: 224.0 (n=64)} \\
\hline
cluster set & big \\
scores & MPVN score=51.92, number of matches with MPD =33\\
keywords & "image model", "real images", "images model", "war", "graphics" \\
centre (sentence 1)& "The model learns the position of the users by learning the complex combination of these components." \\
sentence 2& "The model uses visual representations (spectrograms) of the audio data as input." \\
sentence 3& "The model takes a pair of faces as inputs and performs some form of matching to determine if these two faces belong to the same individual." \\
sentence 4& "The model assesses the targets by choosing the ones that minimize the deviation of the current performances of the observed DMSUs." \\
sentence 5& "The model takes a sequence of disk requests as input and predicts the performance (e.g., average or the 90th percentile) of the workload." \\
\multicolumn{2}{l}{target word: "algorithm", cluster: 257.0 (n=50)} \\
\hline
cluster set & big \\
scores & MPVN score=41.33, number of matches with MPD =12\\
keywords & "dfs tree", "noisy", "refers procedure", "procedure compute", "algorithm refers" \\
centre (sentence 1)& "This algorithm operates on the entire set of statements, recursing by b-prefix depth." \\
sentence 2& "This algorithm essentially computes all the vertices in the next level of the currently built DFS tree simultaneously, building the DFS tree by one level in each pass over the input stream." \\
sentence 3& "This algorithm takes as input an initial set of plans, which at the beginning contains only a single plan." \\
sentence 4& "This algorithm uses the sets V, W and a known upper bound n on the size of the minimal deterministic automaton modeling the black box." \\
sentence 5& "This algorithm looks ahead only when there are no uncovered transitions in the current state and the look ahead is no deeper than the number of steps to the nearest desired state." \\
\multicolumn{2}{l}{target word: "model", cluster: 241.0 (n=22)} \\
\hline
cluster set & MPVN \\
scores & MPVN score=64.54, number of matches with MPD =15\\
keywords & "learning model", "machine learning", "entire population", "population capture", "intentionally inserted" \\
centre (sentence 1)& "Arguably, the machine-learning model can segment the entire population and accommodate individual heterogeneity in an automated way." \\
sentence 2& "In our authentication task, the machine learning model assumes the role of lookup table as described in the previous section." \\
sentence 3& "On subsequent data, the ML model can predict outcomes when presented only with the features." \\
sentence 4& "the deep learning model can still recognize it." \\
sentence 5& "On the other hand, the explanations maybe faithful, however, the machine learning model does not adopt correct evidences to make decisions." \\
\multicolumn{2}{l}{target word: "model", cluster: 73.0 (n=34)} \\
\hline
cluster set & MPVN \\
scores & MPVN score=62.91, number of matches with MPD =7\\
keywords & "utility model", "transmission line", "model relates", "system model", "core model" \\
centre (sentence 1)& "The utility model relates to a black box system of a substation." \\
sentence 2& "The utility model relates to an automobile early-warning device for rear-end collision prevention and reversing and belongs to the field of automobile-mounted electronic devices." \\
sentence 3& "The utility model relates to the technical field of fire protection, and particularly relates to a fire protection supervisory system based on the Internet of things." \\
sentence 4& "The utility model provides an elevator controller with a black box function." \\
sentence 5& "The utility model relates to the field of wireless positioning systems, in particular to an unmanned aerial vehicle system for searching for and locating a black box." \\
\multicolumn{2}{l}{target word: "architecture", cluster: 16.0 (n=22)} \\
\hline
cluster set & MPVN \\
scores & MPVN score=62.7, number of matches with MPD =5\\
keywords & "proposed architecture", "architecture allows", "learning component", "input layer", "architecture enables" \\
centre (sentence 1)& "Furthermore, the proposed architecture allows for a direct incorporation of available human knowledge in form of predefined rules." \\
sentence 2& "Moreover, the proposed architecture divides the overall system to many modules that can be viewed as black boxes with interfaces of inputs and outputs." \\
sentence 3& "The proposed architecture produces a global image representation in a single forward pass." \\
sentence 4& "In addition, the proposed architecture supports a variety of implementation and application options." \\
sentence 5& "Hence, the proposed architecture proves good identification quality." \\
\multicolumn{2}{l}{target word: "model", cluster: 194.0 (n=33)} \\
\hline
cluster set & MPVN \\
scores & MPVN score=61.31, number of matches with MPD =17\\
keywords & "model assumes", "averaging model", "week", "seir model", "model outputs" \\
centre (sentence 1)& "The CPH model incorporates time dependent features." \\
sentence 2& "The SEIR model includes stochasticity in two of its key elements." \\
sentence 3& "The clear box model assumes different parameters with the values outlined in data sheet or are of constant value throughout the system." \\
sentence 4& "The OCC model does not specify how to calculate the thresholds of emotions, but it is believed that they depend on global variables indicating the mood of the agent (Steunebrink et al., 2007b) ." \\
sentence 5& "As is known, the CCR model captures both technical and scale inefficiencies." \\
\multicolumn{2}{l}{target word: "model", cluster: 317.0 (n=20)} \\
\hline
cluster set & MPD \\
scores & MPVN score=39.88, number of matches with MPD =21\\
keywords & "model assign", "model make", "models", "responses model", "points group" \\
centre (sentence 1)& "the model to differentiate between various pairs of attributes" \\
sentence 2& "the model to take advantage of dependencies among the features" \\
sentence 3& "the model to generate suitable translated words" \\
sentence 4& "the model to make better inferences about the content of the signal" \\
sentence 5& "the model to assign weights to windows of text with li le or even no exact query overlap" \\
\multicolumn{2}{l}{target word: "model", cluster: 49.0 (n=22)} \\
\hline
cluster set & MPD \\
scores & MPVN score=47.93, number of matches with MPD =22\\
keywords & "deep neurons", "saliency models", "trained models", "based models", "attacks" \\
centre (sentence 1)& "On the other hand, purely machine-learned models can handle complex dialogs, but they are considered to be black boxes and require large amounts of training data." \\
sentence 2& "Post-hoc models provide much greater applicability as they can be applied to already trained models." \\
sentence 3& "Deep learning-based models have shown the capability to perform mortality prediction, patients subtyping, and diagnosis prediction." \\
sentence 4& "Thirdperson models [10, 18] train additional models from human annotated 'ground truth' reasoning in the form of saliency maps or textual justifications." \\
sentence 5& "Deep models have advanced prediction in many domains, but their lack of interpretability remains a key barrier to the adoption in many real world applications." \\
\multicolumn{2}{l}{target word: "model", cluster: 281.0 (n=26)} \\
\hline
cluster set & MPD \\
scores & MPVN score=38.81, number of matches with MPD =26\\
keywords & "trust model", "proposed", "dialogue model", "state bits", "promoter" \\
centre (sentence 1)& "The proposed TrustE model permeates the trust model with estimates derived from symbolic reasoning, making the act of trust more dynamic and dependent on the agent's history." \\
sentence 2& "This trust model serves as a first example of a model that can be incorporated into the agent's reasoning system." \\
sentence 3& "The TrustE model proposed in this work adds emotions to calculation of trust and reputation for agents." \\
sentence 4& "The multiple indexing and method‐object relations (MIMOR) model tightly integrates a fusion method and a relevance feedback processor into a learning model." \\
sentence 5& "Some trust models [13] also consider norms as input for the trust model and in [26] a method of incorporating this into an MCS is given." \\
\multicolumn{2}{l}{target word: "algorithm", cluster: 283.0 (n=21)} \\
\hline
cluster set & NoMPD \\
scores & MPVN score=34.0, number of matches with MPD =0\\
keywords & "ben tiwari", "interpolating polynomials", "tiwari algorithm", "characteristic", "rings" \\
centre (sentence 1)& "Our algorithm modifies the Ben-Or/Tiwari algorithm from 1988 for interpolating polynomials over rings with characteristic zero to positive characteristics by doing additional probes." \\
sentence 2& "Our algorithm modifies the algorithm of BenOr and Tiwari from 1988 for interpolating polynomials over rings with characteristic zero to characteristic p by doing additional probes." \\
sentence 3& "Unlike Zippel's algorithm and the racing algorithm, our algorithm does not interpolate each variable sequentially and thus can more easily be parallelized." \\
sentence 4& "Our algorithm provides a solution and it is another generalization of the algorithm of Ben-Or and Tiwari." \\
sentence 5& "Our algorithm uses Hensel lifting and extends the EEZ algorithm of Wang which was designed for factorization over" \\
\multicolumn{2}{l}{target word: "algorithm", cluster: 73.0 (n=24)} \\
\hline
cluster set & NoMPD \\
scores & MPVN score=43.13, number of matches with MPD =1\\
keywords & "interval", "ij", "online opponent", "gating algorithm", "conditioning" \\
centre (sentence 1)&  "The separation algorithm computes the . . , pn in the coefficient field K, where c $\in$ K \ {0} is a fixed constant that selects the same associates of the numerator and denominator polynomials for all evaluations." \\
sentence 2& "The forward algorithm computes $\lambda$ t+1 by conditioning $\lambda$ t on x t and applying the transition function p ." \\
sentence 3& "Further, the update algorithm processes every outgoing edge (v, w) of v, i.e., to find vertices w / $\in$ T , which are added to T recursively using the same procedure." \\
sentence 4& "The inversion algorithm uses oracle access to f to invert g f ." \\
sentence 5& "The gating algorithm takes a vote of the final GA generation for the presumed pregame moves (included in the genotype, described in Section III-B)." \\
\multicolumn{2}{l}{target word: "algorithm", cluster: 195.0 (n=20)} \\
\hline
cluster set & NoMPD \\
scores & MPVN score=50.99, number of matches with MPD =1\\
keywords & "quantum algorithm", "smaller needs", "needs make", "metric estimation", "make least" \\
centre (sentence 1)& "Any quantum algorithm that approximates the metric estimation problem with an approximation factor smaller than 3 needs to make at least $\Omega$(n 2 ) oracle calls." \\
sentence 2& "Any quantum algorithm for solving the metric estimation problem with an approximation factor smaller than 3 needs to make at least $\Omega$(n 2 ) oracle calls." \\
sentence 3& "Any randomized classical algorithm that solves STO with bounded probability must use at least $\Omega$(M ) queries to f * ." \\
sentence 4& "Any classical algorithm that makes at most 2 n/6 queries to the oracle finds the exit with probability at most $4 \cdot 2 - n/6$ ." \\
sentence 5& "that any quantum algorithm that approximates metric estimation within a factor smaller than 3, needs to make at least $\Omega$(n 2 ) oracle queries" \\
\multicolumn{2}{l}{target word: "algorithm", cluster: 100.0 (n=21)} \\
\hline
cluster set & NoMPD \\
scores & MPVN score=55.06, number of matches with MPD =2\\
keywords & "test cases", "ml algorithms", "xai algorithms", "onion algorithms", "history information" \\
centre (sentence 1)& "ese algorithms usually use look-up tables to store policies and value functions." \\
sentence 2& "As Table 1 shows, to answer the How question, XAI algorithms commonly employ ranked features, decision trees or rules." \\
sentence 3& "Onion algorithms typically use several layers of encryptions and possibly integrity mechanisms, such as message authentication codes." \\
sentence 4& "Many current XAI algorithms focus on the Why question." \\
sentence 5& "ML algorithms intrinsically consider high-degree interactions between input features, which make disaggregating such functions into human understandable form difficult." \\  
\multicolumn{2}{l}{target word: "model", cluster: 204.0 (n=25)} \\
\hline
cluster set & MPVN \\
scores & MPVN score=61.45, number of matches with MPD =4\\
keywords & "graph", "mis", "quantum", "ranking", "graphs" \\
centre (sentence 1)& "The GCNN model relies on the structure of the graph." \\
sentence 2& "For a random 4-regular graph with 10 nodes, our GGN model has obtained approximately 100\% accuracy in the task of network reconstruction." \\
sentence 3& "The model supports updates in the graph, such that the MIS is not explicitly maintained after each update." \\
sentence 4& "Additionally, the model allows queries of the form In-Mis(v), which reports whether a vertex v $in$ V is present in the MIS of the updated graph." \\
sentence 5& "As shown in Table 1 , the GGN model recovers the ground-truth interaction graph with an accuracy significantly higher than competing method, and the recovery rate in non-chaotic regimes is better than those in chaotic regime." \\
\multicolumn{2}{l}{target word: "algorithm", cluster: 276.0 (n=64)} \\
\hline
cluster set & big \\
scores & MPVN score=41.86, number of matches with MPD =17\\
keywords & "otherwise algorithm", "step algorithm", "request", "smartsearch", "candidate queue" \\
centre (sentence 1)& "At this time, the algorithm uses the Dequeue() operation to extract c together with $\phi$($t 0 t 1 . . . t c+i-1$) from the candidate-queue of interval [i, j] ."  \\
sentence 2& "In this line, the algorithm builds on each one of the intervals [A, x ] and [x , B] a piecewise-linear approximations of $\phi$ (using INDIRECTAPXDEC, IN-DIRECTAPXINC), and stores the corresponding breakpoints in W D , W I , respectively." \\
sentence 3& "If the above condition does not hold, the algorithm starts to process the disk-resident list Lij sequentially." \\
sentence 4& "(9) Otherwise ($\psi$ D (x ) = $\psi$ I (x ) or $\psi$ I (x ) < $\phi$(x )), the algorithm performs a local correction of $\psi$ D and $\psi$ I in lines 9-14, so that a concatenation of $\psi$ D and $\psi$ I is possible, and where (9) still holds." \\
sentence 5& "In line 3, the algorithm checks whether min $\phi$(x) = 0 by performing a call to SMARTSEARCH with a positive query value of less than 1." \\
\multicolumn{2}{l}{target word: "algorithm", cluster: 280.0 (n=27)} \\
\hline
cluster set & NoMPVN \\
scores & MPVN score=30.92, number of matches with MPD =3\\
keywords & "graph algorithm", "pushes flow", "along path", "dfs", "constraint graph" \\
centre (sentence 1)& "Thus, the algorithm finds an s $-$ t path in the residual graph and pushes a flow of one unit along the path." \\
sentence 2& "Hence, on deletion of an edge (x, y) the algorithm first attempts to restore the flow by finding an alternate path from x to y in the residual graph." \\
sentence 3& "Again, the algorithm pushes a flow of unit capacity along the path, restoring the maximum flow in the graph." \\
sentence 4& "If such a path is found the algorithm pushes a flow of unit capacity along the path, restoring the maximum flow in the graph." \\
sentence 5& "Our algorithm computes a DFS tree rooted at r in this augmented graph, where each child subtree of r is a DFS tree of a connected component in the DFS forest of the original graph." \\
\label{tab:neighbours}
\end{longtable}